\documentclass[10pt,twocolumn,letterpaper]{article}

\usepackage[pagenumbers]{cvpr} 

\usepackage[dvipsnames]{xcolor}

\definecolor{cvprblue}{rgb}{0.21,0.49,0.74}
\usepackage[pagebackref,breaklinks,colorlinks,citecolor=cvprblue]{hyperref}
\usepackage{graphicx}
\usepackage{amsmath}
\usepackage{amssymb}
\usepackage{booktabs}

\usepackage[utf8]{inputenc} 
\usepackage[T1]{fontenc}    
\usepackage{url}            
\usepackage{booktabs}       
\usepackage{amsfonts}       
\usepackage{nicefrac}       
\usepackage{microtype}      
\usepackage{xcolor}         
\usepackage{bm}         
\usepackage{amsmath}         
\usepackage{graphicx}

\usepackage{multirow}
\usepackage{makecell}
\usepackage{pifont}
\usepackage{wrapfig}

\usepackage{caption}
\usepackage{subcaption}
\usepackage{graphicx}
\usepackage{comment}
\usepackage{makecell} 
\usepackage{colortbl}

\def\paperID{} 
\def\confName{CVPR}
\def\confYear{2024}

\newcommand{\nickname}{GauHuman}

\title{GauHuman: Articulated Gaussian Splatting from Monocular Human Videos}

\author{
Shoukang Hu\quad\quad
Ziwei Liu\\
S-Lab, Nanyang Technological University\\
\{shoukang.hu, ziwei.liu\}@ntu.edu.sg
}

\begin{document}

\twocolumn[{
    \renewcommand\twocolumn[1][]{#1}%
    \maketitle
    \setlength{\abovecaptionskip}{0.1cm}
    \begin{center}
        \centering
        \includegraphics[width=0.95\textwidth]{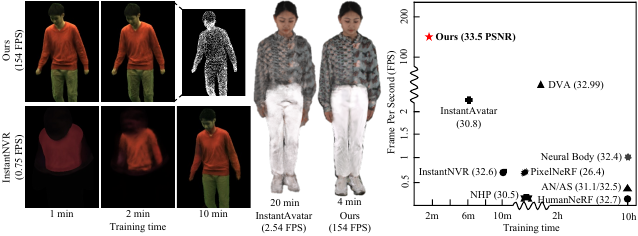}
        \captionof{figure}{\nickname{} models 3D humans with fast training (1$\sim$2 minutes) and real-time rendering ($\sim$189 FPS) given monocular videos. For illustration purposes, we select one sequence (1024p) from MonoCap~\cite{habermann2020deepcap, habermann2021real, peng2022animatable} and one sequence (1024p) from DNA-Rendering~\cite{cheng2023dna}.}
        \label{fig:teaser}
    \end{center}
}]

\maketitle

\begin{abstract}
We present, \textbf{\nickname{}}, a 3D human model with Gaussian Splatting for both \textbf{fast training} (1$\sim$2 minutes) and \textbf{real-time rendering} (up to 189 FPS), compared with existing NeRF-based implicit representation modelling frameworks demanding hours of training and seconds of rendering per frame.
Specifically, \nickname{} encodes Gaussian Splatting in the canonical space and transforms 3D Gaussians from canonical space to posed space with linear blend skinning (LBS), in which effective pose and LBS refinement modules are designed to learn fine details of 3D humans under negligible computational cost. 
Moreover, to enable fast optimization of \nickname{}, we initialize and prune 3D Gaussians with \textbf{3D human prior}, while splitting/cloning via \textbf{KL divergence} guidance, along with a novel \textbf{merge} operation for further speeding up. 
Extensive experiments on ZJU\_Mocap and MonoCap datasets demonstrate that \nickname{} achieves state-of-the-art performance quantitatively and qualitatively with fast training and real-time rendering speed.
Notably, without sacrificing rendering quality, \nickname{} can fast model the 3D human performer with $\sim$13k 3D Gaussians.
Our code is available at \url{https://github.com/skhu101/GauHuman}. 

\end{abstract}    
\section{Introduction}
\label{sec:intro}

Creating high-quality 3D human performers holds extensive utility across various domains, including AR/VR, telepresence, video games, and movie production. 
Recent methods show 3D human avatar can be learned from sparse-view videos~\cite{neuralbody,chen2021animatable, peng2022animatable, xu2021h, noguchi2021neural, weng_humannerf_2022_cvpr, su2021nerf, jiang2022selfrecon, jiang2022neuman, wang2022arah} or even a single image~\cite{weng2023zeroavatar, cha2023generating, liao2023high, huang2023tech} with NeRF-based implicit representation~\cite{mildenhall2021nerf}. 
However, these methods typically require expensive time and computational costs for both training and rendering, which hinders their applications in real-world scenarios.
For example, it normally takes about 10 GPU hours to learn a 3D human performer with a rendering speed of less than 1 frame per second (FPS).

To speed up 3D human modelling, generalizable Human NeRF methods~\cite{kwon2021neural, gao2022mps, choi2022mononhr, zhao2021humannerf, hu2023sherf} trade the quality for less training time. 
They usually fine-tune new human performers with models pre-trained on articulated human data. 
However, such an inefficient \textit{pretrain-and-finetune} paradigm would normally take several hours of pre-training to obtain generalizable representations for 3D humans and an additional one hour of finetuning for each human, while achieving low rendering quality due to the incomplete information within limited conditional input images. On the contrary, a recent line of research focuses on learning efficient 3D human representations~\cite{geng2023learning,jiang2023instantavatar} to improve training speed.
For example, instead of learning neural radiance field (NeRF) from coordinate-based neural networks, methods like \cite{geng2023learning,jiang2023instantavatar} incorporate 3D human modelling with multi-resolution hash encoding (MHE)~\cite{muller2022instant} to accelerate the training.
Although these methods can reduce the training time by a magnitude via human prior like part-based human representations, their rendering speed still limits their applications. 
A critical bottleneck hindering their training and rendering speed arises from the inefficiency of the backward mapping-based ray-cast rendering algorithm employed in these methods, especially at higher resolutions, \eg, 1080p and 4k.

To mitigate this time-consuming dilemma, recent research~\cite{zwicker2001ewa,zhang2022differentiable,kerbl20233d} in point-based rendering can fast render both static and dynamic scenes ~\cite{luiten2023dynamic, yang2023deformable, wu20234d, yang2023real} by splatting a restricted set of 3D points. 
Motivated by these great successes, it is desirable to explore Gaussian Splatting for real-time rendering of articulated objects such as 3D humans with monocular videos. 
Yet, this promising avenue comes with two primary barriers: 1) How to properly integrate articulated human information from monocular videos in the Gaussian Splatting framework? 2) How to efficiently optimize Gaussian Splatting for fast training?
In particular, one potential solution for 3D human modelling with Gaussian Splatting is to directly utilize LBS transformation to transform 3D Gaussians from canonical space into posed space. 
However, it is infeasible to directly apply a 3D vertex point-based LBS transformation for 3D Gaussians, with inaccurate LBS weights and pose information derived from human prior models such as SMPL~\cite{SMPL:2015}.
Meanwhile, current Gaussian Splatting methods utilize either sparse point cloud produced from Structure-from-Motion (SfM)~\cite{schonberger2016structure, snavely2006photo} or random point cloud to initialize 3D Gaussians.
This initialization is specifically designed for static scenes, rather than 3D human structures, making it unsuitable for effective and fast human modelling.
Also, adaptive control of 3D Gaussians proposed in~\cite{kerbl20233d} mainly utilizes gradient information of 3D position to split and clone Gaussians for over-reconstruction and under-reconstruction. Despite improving the rendering performance, it trivially generates a large number of 3D Gaussians, which causes difficulty in optimization and consumes a large storage memory (up to 734 MB in a static scene reconstruction).
More importantly, the Gaussian Splatting framework contains plenty of redundant closely-positioned Gaussians, which holds a potential for additional acceleration. 

To tackle these two challenges, we propose \textbf{\nickname{}}, a 3D human model with articulated Gaussian Splatting representation for both fast training and rendering. 
Our modelling framework and fast optimization algorithm are detailed as follows. 
\textbf{Articulated Gaussian Splatting with integrated human information}.
Inspired by previous Human NeRF methods, we encode Gaussian Splatting in a canonical space and wrap it to a posed space. 
Thanks to the Gaussian property, we can transform 3D Gaussians from a canonical space to a posed space via linear blend skinning (LBS). 
We rotate and translate the 3D position and covariance of each 3D Gaussian with the estimated LBS transformation matrix.
However, directly applying multi-layer perceptron (MLP) to estimate LBS weights would be time-consuming and generate low-quality rendering results. 
To address this, we start with the LBS weights from SMPL~\cite{SMPL:2015} and employ an MLP to predict corresponding LBS weight offsets, followed by an effective pose refinement module to rectify SMPL poses for more accurate LBS transformation.
For rendering an image with 1024p, our \nickname{} only needs to splat a limited set (\eg, 13k) of 3D Gaussians, in comparison with Human NeRF methods that apply volume rendering to millions of sampled points in the whole 3D space. 
This significantly reduces the computational burden in both training and rendering processes.
\textbf{Fast Optimization of 3D Gaussian Splatting}.
1) \textit{3D Gaussian Initialization}. 
In \nickname{}, we utilize the human prior like SMPL for 3D Gaussian initialization, which enables fast optimization of articulated Gaussian Splatting as the initialization encodes human structures. 
2) \textit{Split/Clone/Merge}.
Our analysis indicates that the splitting and cloning process overlooks a crucial metric, namely, the distance between 3D Gaussians, during the split and clone process.
In light of this, we integrate the Kullback–Leibler (KL) divergence of 3D Gaussians to regulate the split and clone process.
A novel merge operator is also proposed to merge 3D Gaussians that are close to each other for further speeding up.
3) \textit{Prune}. As 3D human has specific structures (human prior), we prune 3D Gaussians that are far away from the human body, \eg, SMPL vertices. 
With the above algorithm implemented in articulated 3D Gaussian Splatting, we can fast model 3D human performers with $\sim$13k 3D Gaussians (3.5MB memory storage) within one or two minutes. 
Our main contributions are as follows:
\begin{enumerate}
    \item  We present, \nickname{}, a 3D model with articulated Gaussian Splatting for both fast training (1$\sim$2 minutes) and real-time rendering (up to 189 FPS), in which pose and LBS refinement modules are designed to learn fine details of 3D humans. 
    \item To enable fast optimization of articulated 3D Gaussian Splatting, we initialize and prune 3D Gaussians with 3D human prior, while splitting/cloning via KL divergence measures, accompanied by a novel merge operation for further speeding up.
    \item Experiments on two monocular data sets validate that our \nickname{} achieves state-of-the-art novel view synthesis performance with fast training and real-time rendering.
\end{enumerate}

\section{Related Work}
\label{sec:related_work}

\noindent \textbf{Rendering and Radiance Fields.}
Early novel view synthesis approaches focus on using light fields for densely sampled~\cite{gortler2023lumigraph, levoy2023light} and unstructured~\cite{buehler2023unstructured} capture.  
In later years, the advent of Structure-from-Motion (SfM)~\cite{snavely2006photo} enables a new domain by synthesizing novel views from a collection of photos with blending techniques~\cite{goesele2007multi, chaurasia2013depth, eisemann2008floating, hedman2018deep, kopanas2021point, tewari2022advances}. 
With the development of deep learning techniques, previous research~\cite{flynn2016deepstereo, zhou2016view, hedman2018deep, riegler2020free, thies2019deferred} propose to adopt deep learning techniques with multi-view stereo (MVS) for novel view synthesis.
Recently, deep learning techniques have been used to learn implicit neural radiance fields~\cite{penner2017soft, henzler2019escaping, sitzmann2019deepvoxels, mildenhall2021nerf} with volume rendering~\cite{kajiya1984ray, max1995optical}. 
NeRF~\cite{mildenhall2021nerf} is one of the representative work, learning high-quality novel view synthesis results, which has inspired a series of follow-up works~\cite{barron2022mip, yu2021pixelnerf, wang2021neus, muller2022instant, chibane2021stereo, trevithick2021grf, xu2022point, wang2021ibrnet, yang2020cost, liu2022neural, chen2021mvsnerf, jang2021codenerf, rematas2021sharf, darmon2022improving, deng2021depth, johari2022geonerf, jain2021putting, niemeyer2021regnerf, zhang2020nerf++}.
To accelerate training and rendering speed, several methods~\cite{chen2022tensorf, garbin2021fastnerf, li2022nerfacc, liu2020neural, muller2022instant, reiser2021kilonerf, fridovich2022plenoxels, sun2022direct, takikawa2021neural, yu2021plenoctrees} are proposed, including replacing multi-layer perceptrons (MLPs) with more efficient representations or skipping the empty space via the occupancy grid.
In contrast to NeRF, which applies volume rendering to millions of sampled points in the whole 3D space when rendering an image, point-based rendering~\cite{gross2011point, grossman1998point, sainz2004point, aliev2020neural, ruckert2022adop, lassner2021pulsar, kopanas2022neural, zhang2022differentiable} splats a restricted set of points. 
This approach provides flexibility and efficiency in reconstructing 3D scenes by optimizing the opacity and position of individual points.
Furthermore, point primitives can be represented as circular, elliptic
discs, ellipsoids or surfels for high quality results~\cite{botsch2005high, pfister2000surfels, ren2002object, zwicker2001surface, wiles2020synsin, yifan2019differentiable}. 
To achieve real-time rendering of large-scale scenes, 3D Gaussians with splatting techniques are used to represent static scenes\cite{kerbl20233d}, dynamic scenes~\cite{luiten2023dynamic, yang2023deformable, wu20234d, yang2023real}, and other tasks~\cite{stoll2011fast, rhodin2015versatile, wang2022voge}.
In this work, we further extend Gaussian Splatting to articulated 3D humans with both fast training and real-time rendering.

\noindent \textbf{3D Human Modelling.}
Although 2D human modelling~\cite{albahar2021pose, lewis2021tryongan, sarkar2021humangan, men2020controllable, fu2022stylegan, jiang2022text2human, jiang2023text2performer, fu2023unitedhuman} has made substantial progress, it is still challenging to reconstruct high-quality 3D humans for synthesizing free-viewpoint videos.
Traditional methods either leverage multi-view stereo methods~\cite{guo2019relightables, schonberger2016pixelwise} or depth fusion~\cite{collet2015high, dou2016fusion4d, su2020robustfusion} to reconstruct 3D human geometry, which typically requires dense camera arrays or depth sensors. 
To alleviate this, previous research~\cite{alldieck2022photorealistic, saito2019pifu, saito2020pifuhd, kwon2021neural, gao2022mps, choi2022mononhr, zhao2021humannerf, hu2023sherf} learn human priors from data sets containing either 3D ground truth human models or multi-view image collections, which enables them to synthesize novel views even with a single image.
However, due to the limited diversity in pre-training data sets, these methods can not generalize well to 3D humans with complex poses.
Another line of works reconstruct 3D human performers with implicit neural representations from sparse-view videos~\cite{neuralbody,chen2021animatable, peng2022animatable, peng2022animatable, xu2021h, noguchi2021neural, weng_humannerf_2022_cvpr, su2021nerf, jiang2022selfrecon, jiang2022neuman, wang2022arah, liu2021neural, lin2022efficient} or even a single image~\cite{weng2023zeroavatar, cha2023generating, liao2023high, huang2023tech}.
However, these methods usually require hours of training time and seconds of rendering time per frame, restricting their applications to real-world scenarios.
To improve the training efficiency, recent research~\cite{geng2023learning, jiang2023instantavatar} apply multi-hashing encoding (MHE) to reduce the training time to several minutes, while their rendering remains comparatively slow.
Moreover, neural volumetric primitives~\cite{lombardi2021mixture, chen2023primdiffusion, remelli2022drivable} have been proposed for efficient rendering of 3D humans, but they still require hours of training time.
In this work, we focus on reconstructing high-quality 3D humans from monocular videos with both fast training and real-time rendering.
Our concurrent work~\cite{xu20234k4d} learns real-time rendering of dynamic 3D humans from dense-view videos.

\section{Our Approach}
\label{sec:method}

The goal of this paper is to learn 3D human performers with fast training and rendering from monocular videos. 
We assume the calibrated camera parameters and the human region masks are known. 
We also assume the corresponding SMPL parameters $\bm{\beta}, \bm{\theta}$ are given.

\subsection{Preliminary}
\label{sec:preliminary}
\noindent \textbf{SMPL}~\cite{SMPL:2015} is a parametric human model $M(\bm{\beta}, \bm{\theta})$, which defines $\bm{\beta}, \bm{\theta}$ to control body shapes and poses. 
In this work, we apply the Linear Blend Skinning (LBS) algorithm used in SMPL to transform points from a canonical space to a posed space. 
For example, 3D point $\bm{p}^c$ in the canonical space is transformed to a posed space defined by pose $\bm{\theta}$ as $\bm{p}^t = \sum_{k=1}^{K}w_{k}(\bm{G_k}(\bm{J}, \bm{\theta})\bm{p}^c+\bm{b_k}(\bm{J}, \bm{\theta}, \bm{\beta}))$, where $\bm{J}$ includes $K$ joint locations, $\bm{G_k}(\bm{J}, \bm{\theta})$ and $\bm{b_k}(\bm{J}, \bm{\theta}, \bm{\beta})$ are the transformation matrix and translation vector of joint $k$ respectively, $w_{k}$ is the linear blend weight. 

\begin{figure*}[t]
    \centering
    \vspace{-3mm}
    \includegraphics[width=16cm]{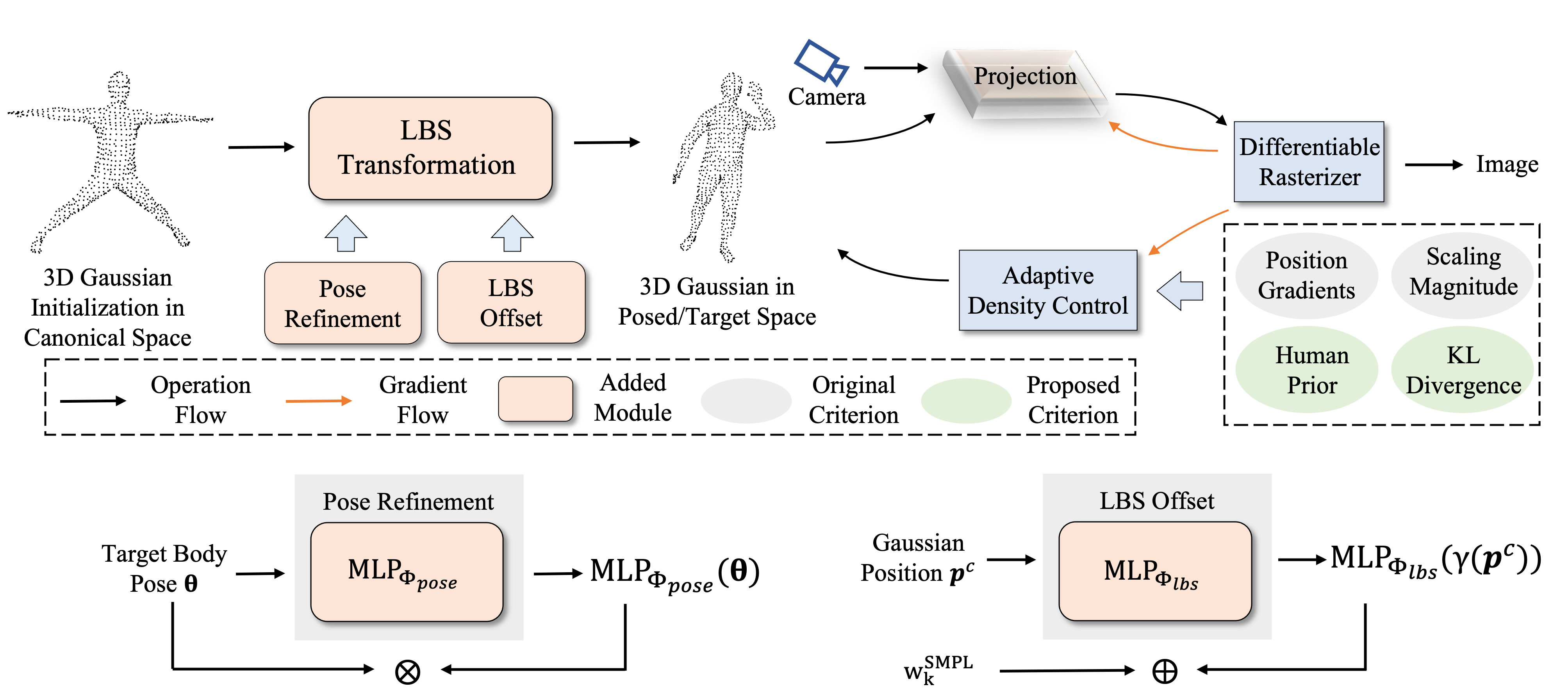}
    \setlength{\abovecaptionskip}{0cm}
    \caption{\textbf{\nickname{} Framework.} We first initialize 3D Gaussians' positions $\bm{p}^{c}$ from SMPL vertex points. Then we incorporate pose refinement module $\text{MLP}_{\Phi_\text{pose}}(\cdot)$ and LBS weight field $\text{MLP}_{\Phi_\text{lbs}}(\cdot)$ module to learn the LBS transformation to transform 3D Gaussians from canonical space to posed space. During optimization, a tile-based differentiable rasterizer is applied to enable fast rendering. To adaptively control the number of 3D Gaussians, we further propose to use human prior (\eg, SMPL) and KL divergence measure to regulate the split, clone, merge, and prune process.} 
\label{fig: overview}
\vspace{-4mm}
\end{figure*}

\noindent \textbf{3D Gaussian Splatting}~\cite{kerbl20233d}
In comparison with NeRF methods that learn continuous structured volumetric representation for the whole 3D space, Gaussian Splatting represents a 3D scene with a set of unstructured discrete 3D Gaussians.  
Each 3D Gaussian is defined with a point center $\bm{p}$ and a full 3D covariance matrix $\bm{\Sigma}$ in a world space, i.e., 
\begin{equation}
\label{eqn:3d_gaussian_splatting}
\setlength{\abovedisplayskip}{5pt} 
\setlength{\belowdisplayskip}{5pt}
\begin{aligned}
G(\bm{x})=\frac{1}{(2\pi)^{\frac{3}{2}}|\bm{\Sigma}|^{\frac{1}{2}}}e^{-\frac{1}{2}(\bm{x}-\bm{p})^{T}\bm{\Sigma}^{-1}(\bm{x}-\bm{p})}.
\end{aligned}
\end{equation}

To ensure the positive semi-definiteness of 3D covariance matrix $\bm{\Sigma}$, $\bm{\Sigma}$ is decomposed into two learnable components, \ie, a quaternion $\bm{r}\in\mathcal{R}^{4}$ representing rotation and a vector $\bm{s}\in\mathcal{R}^{3}$ representing scaling.
Through the transformation of quaternion $\bm{r}$ into a rotation matrix $\bm{R}$ and scaling vector $\bm{s}$ into a scaling matrix $\bm{S}$, the associated covariance matrix $\bm{\Sigma}$ is defined as $\bm{\Sigma}=\bm{R}\bm{S}\bm{S}^{T}\bm{R}^{T}$.

To splat 3D Gaussians, 3D Gaussian Splatting~\cite{zwicker2001ewa, zwicker2001surface} first transforms them from world space to ray coordinate and then projects them to a 2D screen plane.
As shown in~\cite{zwicker2001ewa}, utilizing a world-to-camera transformation matrix $\bm{W}$, translation vector $\bm{d}$, and the Jacobian matrix $\bm{J}$ representing an approximated projective transformation from camera coordinate to ray coordinate, 3D Gaussians in ray coordinate assume the following form:
\begin{equation}
\setlength{\abovedisplayskip}{5pt} 
\setlength{\belowdisplayskip}{5pt}
\begin{aligned}
G(\bm{x})=\frac{1}{(2\pi)^{\frac{3}{2}}|\bm{\Sigma}^{'}|^{\frac{1}{2}}}e^{-\frac{1}{2}(\bm{x}-\bm{p}^{'})^{T}{\bm{\Sigma}^{'}}^{-1}(\bm{x}-\bm{p}^{'})}
\end{aligned}
\label{eqn:gs_world_to_ray}
\end{equation}
where $\bm{p}^{'}=\bm{m}(\bm{W}\bm{p}+\bm{d})$ and $\bm{\Sigma}^{'}=\bm{J}\bm{W}\bm{\Sigma}\bm{W}^{T}\bm{J}^{T}$ are the mean vector and covariance matrix of 3D Gaussians in ray coordinate,
$\bm{m}(\cdot)$ denotes the mapping from camera coordinate to ray coordinate. 
The projected 2D Gaussians in the 2D screen plane can then be obtained by integrating Eqn.~(\ref{eqn:gs_world_to_ray}) along the projection direction or utilizing a property mentioned by~\cite{zwicker2001ewa}.

After mapping 3D Gaussians to a 2D screen space, we count 2D Gaussians that overlap with each pixel and calculate their color $c_i$ and density $\alpha_i$ contribution.
Finally, the colors of each pixel can be obtained by blending $N$ ordered Gaussians:
\begin{equation}
\setlength{\abovedisplayskip}{5pt} 
\setlength{\belowdisplayskip}{5pt}
\begin{aligned}
\hat{C}=\sum_{i\in N} c_i\alpha_i\prod_{j=1}^{i-1}(1-\alpha_i).
\end{aligned}
\end{equation}

During optimization, 3D Gaussians are first initialized from either Structure-from-Motion (SfM) or random sampling. 
Then adaptive control of Gaussians is employed to improve the rendering quality, which mainly includes three operations, \ie, split, clone, and prune. 
The split and clone operations are performed on 3D Gaussians with large position gradients.
These 3D Gaussians would be 1) split into smaller Gaussians if the magnitude of the scaling matrix is larger than a threshold or 2) cloned if the magnitude of the scaling matrix is smaller than a threshold.
While the splitting and cloning process augments the number of Gaussians, the prune operation selectively eliminates Gaussians with either excessively small opacity ($\alpha$) or overly large scaling magnitudes.

\subsection{Articulated 3D Gaussian Splatting}
Recent research can fast render static scenes~\cite{kerbl20233d} or dynamic scenes~\cite{luiten2023dynamic, yang2023deformable, wu20234d, yang2023real} by
splatting a restricted set of 3D points. 
Motivated by these great successes, it is desirable to explore Gaussian Splatting for real-time rendering of articulated objects such as 3D humans with monocular videos.
When employing Gaussian Splatting to articulated 3D humans, there are two main challenges, \ie, 1) How to properly integrate articulated human information from monocular videos in the Gaussian Splatting framework? ii). how to efficiently optimize Gaussian Splatting for fast training?

Inspired by previous Human NeRF research, as shown in Fig.~\ref{fig: overview}, we represent 3D humans with 3D Gaussians in a canonical space and then wrap 3D Gaussians from the canonical space to posed space through forward LBS transformation. 
Thanks to the Gaussian property, we can transform 3D Gaussians from a canonical space to a posed space via linear blend skinning (LBS). 
Specifically, we rotate and translate the 3D position and covariance of each 3D Gaussian with the estimated LBS transformation matrix.
\begin{equation}
\begin{aligned}
\bm{p}^{t}&= \bm{G}(\bm{J}^t, \bm{\theta}^t)\bm{p}^c+\bm{b}(\bm{J}^t, \bm{\theta}^t, \bm{\beta}^t) \\
\bm{\Sigma}^{t}&=\bm{G}(\bm{J}^t, \bm{\theta}^t)\bm{\Sigma}^c\bm{G}(\bm{J}^t, \bm{\theta}^t)^{T},
\end{aligned}
\end{equation}
where $\bm{p}^{t}$, $\bm{\Sigma}^{t}$, $\bm{p}^{c}$, and $\bm{\Sigma}^{c}$ are the 3D position vector and covariance matrix in the canonical and posed space respectively. $\bm{G}(\bm{J}^t, \bm{\theta}^t)=\sum_{k=1}^{K}w_{k}\bm{G_k}(\bm{J}^t, \bm{\theta}^t)$, $\bm{b}(\bm{J}^t, \bm{\theta}^t, \bm{\beta}^t)=\sum_{k=1}^{K}w_{k}\bm{b_k}(\bm{J}^t, \bm{\theta}^t, \bm{\beta}^t)$ are the rotation matrix and translation vector. $K$ is the joint number, $\bm{G_k}(\bm{J}^t, \bm{\theta}^t)$ and $\bm{b_k}(\bm{J}^t, \bm{\theta}^t, \bm{\beta}^t)$ are the transformation matrix and translation vector of joint $k$ respectively, $w_{k}$ is the linear blend skinning (LBS) weight we need to estimate.

\noindent \textbf{LBS Weight Field} Multi-layer perceptrons (MLPs) can be used to predict the LBS weight coefficients $w_k$ for each 3D Gaussian in the canonical space. 
However, it would be time-consuming and generate low-quality rendering results. 
To address this, we start with the LBS weights from  SMPL and employ an $\text{MLP}_{\Phi_\text{lbs}}(\cdot)$ to predict corresponding LBS weight offsets. 
Specifically, for each 3D Gaussian, we add the LBS weight $w_k^{\text{SMPL}}$ of nearest SMPL vertex with the predicted offsets $\text{MLP}_{\Phi_\text{lbs}}(\gamma(\bm{p}^{c}))$, \ie, 
\begin{equation}
\begin{aligned}
w_k=\frac{e^{\log(w_k^{\text{SMPL}}+10^{-8})+\text{MLP}_{\Phi_\text{lbs}}(\gamma(\bm{p}^{c}))[k]}}{\sum_{k=1}^{K}e^{\log(w_k^{\text{SMPL}}+10^{-8})+\text{MLP}_{\Phi_\text{lbs}}(\gamma(\bm{p}^{c}))[k]}},
\end{aligned}
\end{equation} 
where we apply the standard positional encoding $\gamma(\cdot)$ to the input (3D postion $\bm{p}^c$) of $\text{MLP}_{\Phi_\text{lbs}}(\cdot)$.
The idea of learning LBS weight field is also applied in previous human modelling works~\cite{bhatnagar2020loopreg,chen2021snarf,deng2020nasa,huang2020arch,mihajlovic2021leap,peng2021animatable,saito2021scanimate,tiwari2021neural,weng2020vid2actor,yang2021s3}.

\noindent \textbf{Pose Refinement}
The body poses $\bm{\theta}$ estimated from images is often inaccurate, which degrades the rendering quality. 
To address this, we incorporate a pose correction module $\text{MLP}_{\Phi_\text{pose}}(\cdot)$ to learn relative updates to joint angles:
\begin{equation}
\setlength{\abovedisplayskip}{5pt} 
\setlength{\belowdisplayskip}{5pt}
\begin{aligned}
\bm{\theta}=\bm{\theta}^{\text{SMPL}}\otimes\text{MLP}_{\Phi_\text{pose}}(\bm{\theta}),
\end{aligned}
\end{equation} 
where $\bm{\theta}^{\text{SMPL}}$ is the SMPL body pose parameters estimated from images. 
The body pose refinement is also investigated in~\cite{weng_humannerf_2022_cvpr, jiang2023instantavatar, chen2021animatable}. 

During inference, thanks to the 3D Gaussian-based representation defined in a canonical space, the learned LBS weight for each 3D Gaussian can be derived from the LBS weight field module in advance, which saves computation time.
Meanwhile, the rectified pose parameters can also be saved from the pose refinement module at the end of training. 
Considering the above, our \nickname{} can model high-quality 3D humans with high rendering speed.

\begin{table*}[t]
\setlength{\abovecaptionskip}{0cm}
\caption{Quantitative comparison of our \nickname{} and baseline methods on the ZJU\_MoCap and MonoCap data sets. We use bold and underlined text to denote the best and second-best results of each metric respectively. Our \nickname{} achieves the SOTA PSNR and LPIPS with fast training (1$\sim$2 minutes) and real-time rendering (up to 189 FPS) in both data sets. Note that PixelNeRF~\cite{yu2021pixelnerf} and NHP~\cite{kwon2021neural} are pre-trained for an additional 10 hours. $^\dag$ denotes the fine-tuning time for generalizable methods. LPIPS$^*$ = 1000 $\times$ LPIPS. Frames per second (FPS) is measured on an RTX 3090. \textbf{For a fair comparison, we do not conduct test-time optimization of SMPL parameters with images from the test set on InstantAvatar~\cite{jiang2023instantavatar}.}
}
\centering
\label{tab: main_result}
\begin{tabular}{l|ccccc|ccccc}
\toprule
\multirow{2}*{Method} & \multicolumn{5}{c|}{ZJU\_MoCap} & \multicolumn{5}{c}{MonoCap} \\
~ & PSNR$\uparrow$ & SSIM$\uparrow$ & LPIPS$^*$$\downarrow$ & Train & FPS & PSNR$\uparrow$ & SSIM$\uparrow$ & LPIPS$^*$$\downarrow$ & Train & FPS \\
\midrule
PixelNeRF~\cite{yu2021pixelnerf} & 24.71 & 0.892 & 121.86 & 1h$^\dag$ & 1.20 & 26.43 & 0.960 & 43.98 & 1h$^\dag$ & 0.75 \\
NHP~\cite{kwon2021neural} & 28.25 & 0.955 & 64.77 & 1h$^\dag$ & 0.15 & 30.51 & 0.980 & 27.14 & 1h$^\dag$ & 0.05 \\
NB~\cite{neuralbody} & 29.03 & 0.964 & 42.47 & 10h & 1.48 & 32.36 & 0.986 & 16.70 & 10h & 0.98 \\
AN~\cite{peng2021animatable} & 29.77 & 0.965 & 46.89 & 10h & 1.11 & 31.07 & 0.985 & 19.47 & 10h & 0.31\\
AS~\cite{peng2022animatable} & 30.38 & \textbf{0.975} & 37.23 & 10h & 0.40 & 32.48 & \textbf{0.988} & \textbf{13.18} & 10h & 0.29 \\
HumanNeRF~\cite{weng_humannerf_2022_cvpr} & 30.66 & 0.969 & \underline{33.38} & 10h & 0.30 & \underline{32.68} & \underline{0.987} & 15.52 & 10h & 0.08 \\
DVA~\cite{remelli2022drivable} & 29.45 & 0.956 & 37.74
& 1.5h & \underline{16.5} & \underline{32.99} & 0.983 & 15.83 & 1.5h &\underline{10.5} \\
InstantNVR~\cite{geng2023learning} &  \underline{31.01} & \underline{0.971} & 38.45 & 5m & 2.20 & 32.61 & \textbf{0.988} & 16.68 & 10m & 0.75 \\
InstantAvatar~\cite{jiang2023instantavatar} &  29.73 & 0.938 & 68.41 & \underline{3m} & \underline{4.15} & 30.79 & 0.964 & 39.75 & \underline{6m} & \underline{2.54} \\
\textbf{\nickname{}}(Ours) & \textbf{31.34} & 0.965 & \textbf{30.51} & \textbf{1m} & \textbf{189} & \textbf{33.45} & 0.985 & \underline{13.35} & \textbf{2m} & \textbf{154} \\
\bottomrule
\end{tabular}
\vspace{-4mm}
\end{table*}

\subsection{Fast Optimization of \nickname{}}
\label{sec: efficient_opt}
The current optimization scheme of Gaussian Splatting~\cite{kerbl20233d} is specifically designed for static scenes. 
To facilitate fast optimization for articulated 3D humans, we incorporate 3D human prior and KL divergence measures to speed up.

\noindent \textbf{3D Gaussian Initialization}
Current Gaussian Splatting methods utilize either Structure-from-Motion (SfM)~\cite{schonberger2016structure, snavely2006photo} or random point cloud to initialize 3D Gaussians, which overlooks 3D human structures. 
In \nickname{}, we initialize 3D Gaussians with human prior like SMPL vertex points. 
The above initialization encodes human structure, thus accelerating optimization. 

\noindent \textbf{Split \& Clone \& Merge \& Prune}
As mentioned in Sec.~\ref{sec:preliminary}, the optimization of Gaussian Splatting relies on the adaptive control of 3D Gaussians to ensure superior rendering quality.
To adaptively control the number of Gaussians, three operations, i.e., split, clone, and prune, are performed based on the position gradients and magnitude of the scaling matrix. 
However, although split and clone operations substantially improve the performance, they trivially generate a substantial quantity of 3D Gaussians. 
This results in a slowdown of the optimization and a significant consumption of storage memory, reaching up to 734 MB in static scenes.
Our analysis indicates that previous research overlooks a crucial metric, namely, the distance between 3D Gaussians, during the split and clone process. 
For example, if two Gaussians are close to each other, even if the position gradients are larger than a threshold, they should not be split or cloned as 1) these Gaussians are updating their positions to improve their performance, 2) splitting and cloning these Gaussians have negligible influence on the final performance as they are too close to each other. 
To regulate the split and clone process, we propose to use Kullback–Leibler (KL) divergence as a measure of the distance of 3D Gaussians, \ie,
\begin{equation}
\setlength{\abovedisplayskip}{5pt} 
\setlength{\belowdisplayskip}{5pt}
\begin{aligned}
KL(G(\bm{x}_0)|G(\bm{x}_1&))=
\frac{1}{2}(tr(\bm{\Sigma}_1^{-1}\bm{\Sigma}_0)+\ln\frac{\det \bm{\Sigma}_1}{\det \bm{\Sigma}_2}\\
&+(\bm{p}_1-\bm{p}_0)^{T}\bm{\Sigma}_1^{-1}(\bm{p}_1-\bm{p}_0)-3),
\end{aligned}
\label{eqn:kl_div}
\end{equation} 
where $\bm{p}_0$, $\bm{\Sigma}_0$, $\bm{p}_1$,  $\bm{\Sigma}_1$ are the position and covariance matrix of two 3D Gaussians $G(\bm{x}_0)$ and $G(\bm{x}_1)$.

After obtaining the KL divergence of nearby 3D Gaussian pairs, we select 3D Gaussians with large KL divergence and positional gradients to perform the split and clone.
However, the above requires calculating the KL divergence for each pair of 3D Gaussians, which consumes lots of time during the optimization.
To solve this problem, we further adopt two approaches to simplify the whole calculation. 
1) For each 3D Gaussian, we first identify its closest 3D Gaussian through the widely adopted k-nearest neighbor (k-NN) algorithm, which assesses the distance between their respective centers.
Then we calculate a KL divergence for each pair of nearby 3D Gaussians. 
The aforementioned simplification effectively reduces the time complexity from $O(U^2)$ to $O(U)$, where $U$ represents the total number of 3D Gaussians.
2) As the covariance matrix is decomposed into the product of rotation and scaling matrices, we further simplify the computation of matrix inverse and determinant operations in Eqn.~(\ref{eqn:kl_div}) with the diagonal and orthogonal property of rotation and scaling matrices.
Details are shown in the appendix. 

In addition to split and clone operations, we further propose a novel merge operation, which merges redundant 3D Gaussians with 1) large position gradients, 2) small scaling magnitude, and 3) KL divergence less than 0.1. 
Specifically, we merge two Gaussians by averaging their positions, opacity, and SH coefficients. Additionally, we scale the covariance (scaling) matrix of the first Gaussian by a factor of 1.25 as the initialization for the new Gaussian. 
For prune operation, we incorporate 3D human prior like SMPL to prune 3D Gaussians far away from SMPL vertex points.  

During optimization, four loss functions are used, \ie,\\
\begin{equation}
\setlength{\abovedisplayskip}{5pt} 
\setlength{\belowdisplayskip}{5pt}
\label{loss: overall}
\begin{aligned}
\mathcal{L} = \mathcal{L}_{color} + \lambda_{1}\mathcal{L}_{mask} + \lambda_{2}\mathcal{L}_{SSIM} + \lambda_{3}\mathcal{L}_{LPIPS},
\end{aligned}
\end{equation}
where $\lambda$'s are loss weights. Empirically, we set $\lambda_1=0.5$, $\lambda_2 = \lambda_3 = 0.01$ to ensure the same magnitude for each loss. Details are shown in the appendix.

\section{Experiments}
\label{sec:exp}

\subsection{Experimental Setup}
\noindent \textbf{Datasets.} 
Experiments are conducted on two widely used human modelling data sets, \ie, ZJU\_MoCap~\cite{neuralbody} and MonoCap~\cite{peng2022animatable, habermann2020deepcap, habermann2021real}. 
For ZJU\_MoCap, same as previous work~\cite{weng_humannerf_2022_cvpr, geng2023learning}, we select 6 human subjects (377, 386, 387, 392, 393, 394) to conduct experiments.
We also adopt the same training and testing setting as~\cite{geng2023learning}, \ie, one camera is used for training, while the remaining cameras are used for evaluation. 
For each subject, we sample 1 frame every 5 frames and collect 100 frames for training.
The foreground masks, camera, and SMPL parameters provided by the data set are used for evaluation purposes.
MonoCap data set contains four multi-view videos collected by~\cite{peng2022animatable} from the DeepCap data set~\cite{habermann2020deepcap} and the DynaCap data set~\cite{habermann2021real}.
Similar to ZJU\_MoCap, it provides human masks, cameras, and SMPL parameters.
We adopt the same camera views as~\cite{peng2022animatable} for training and testing.
For each subject, we collect 100 frames for training by sampling 1 frame every 5 frames.
Implementation details and evaluation metrics are shown in the appendix.

\noindent \textbf{Comparison Methods.} 
We compare our \nickname{} with two categories of state-of-the-art human modelling methods, \ie, subject-specific optimization based methods~\cite{neuralbody, peng2021animatable, peng2022animatable, weng_humannerf_2022_cvpr, geng2023learning, jiang2023instantavatar, remelli2022drivable} and generalizable methods~\cite{yu2021pixelnerf, kwon2021neural}. 
Following~\cite{geng2023learning}, we report the average metric values of all selected human subjects.
For a fair comparison, we do not conduct test-time optimization of SMPL parameters with images from the test set on InstantAvatar~\cite{jiang2023instantavatar}.
More details are shown in the appendix.

\begin{figure*}[t]
    \vspace{-3mm}
    \setlength{\abovecaptionskip}{0cm}
    \centering
    \includegraphics[width=17cm]{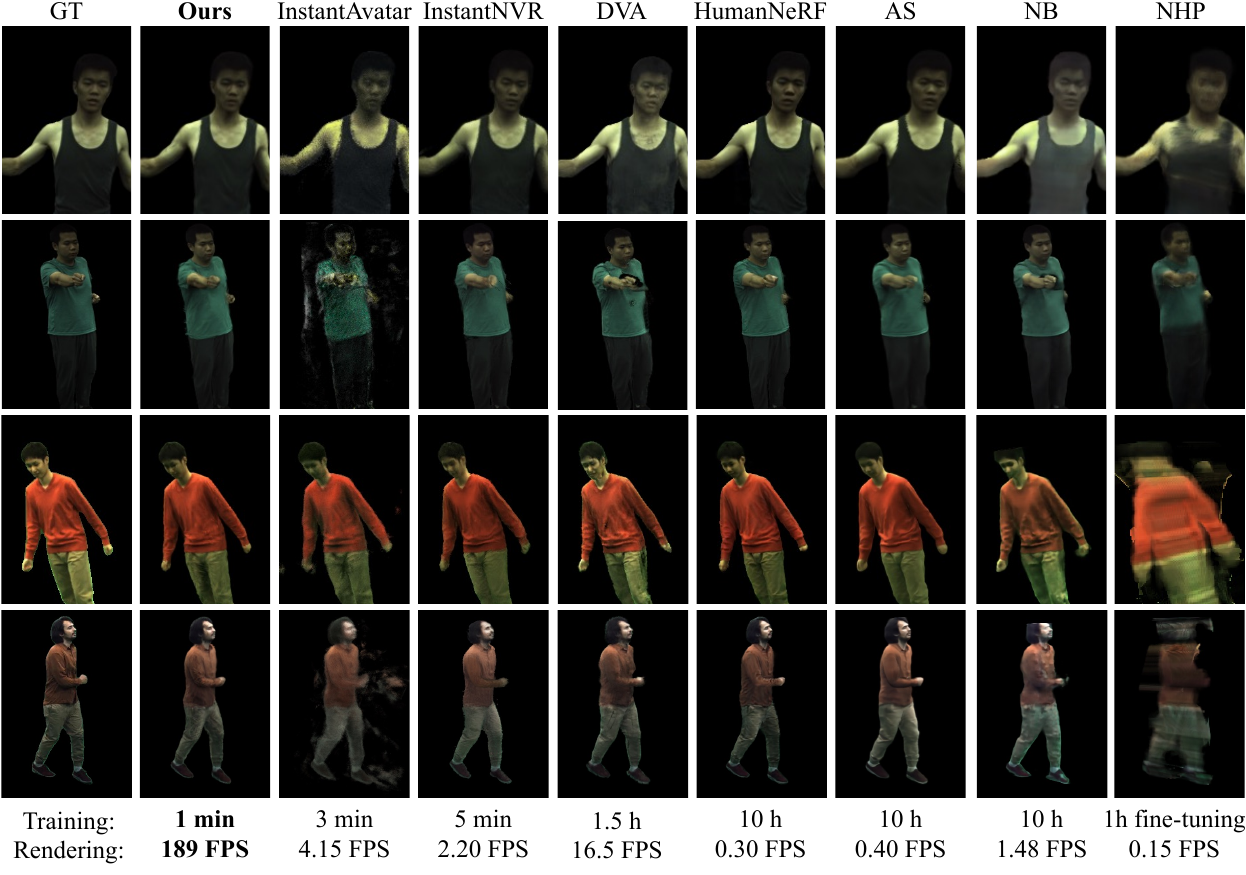}
    \setlength{\abovecaptionskip}{0cm}
    \caption{Novel view synthesis results produced by our \nickname{} and baseline methods on ZJU\_MoCap and MonoCap. The bottom lines show the training time and rendering speed of each method on ZJU\_Mocap. Zoom in for the best view.} 
\label{fig: main_vis}
\vspace{-3mm}
\end{figure*}

\subsection{Quantitative Results}
Tab.~\ref{tab: main_result} summarizes the overall novel view synthesis results of our \nickname{}, NB~\cite{neuralbody}, AN~\cite{peng2021animatable}, AS~\cite{peng2022animatable}, HumanNeRF~\cite{weng_humannerf_2022_cvpr}, DVA~\cite{remelli2022drivable}, InstantNVR~\cite{geng2023learning}, InstantAvatar~\cite{jiang2023instantavatar}, PixelNeRF~\cite{yu2021pixelnerf}, and NHP~\cite{kwon2021neural}.
Thanks to our proposed articulated 3D human Gaussian-based representation, we can fast optimize it for around 1 or 2 minutes to achieve photorealistic rendering results, while baseline methods (\ie, NB, AN, AS, and HumanNeRF) take around 10 hours to finish the training and generalizable methods takes about 1 hour for fine-tuning after 10 hours' pre-training.
To accelerate the training of 3D human modelling, InstantNVR~\cite{geng2023learning} and InstantAvatar~\cite{jiang2023instantavatar} reduce the training to several minutes with multi-hashing encoding representation.
However, their rendering speed still restricts their applications in real-world applications.
When compared with these two state-of-the-art methods for fasting training, our \nickname{} achieves slightly better performance with 3 times faster training and 45 times faster rendering speed.
Among baseline methods, DVA achieves a fast rendering speed (up to 16.5 FPS) with efficient mixtures of volumetric primitives.
However, in contrast to our \nickname{}, DVA takes about 90x more time to converge and 11.45x more time to render one image.

\subsection{Qualitative Results}
As shown in Fig.~\ref{fig: main_vis}, our \nickname{} shows better appearance and geometry details than DVA~\cite{remelli2022drivable},  NB~\cite{neuralbody} and NHP~\cite{kwon2021neural}. 
Although NB~\cite{neuralbody} and NHP~\cite{kwon2021neural} show impressive results with 4-view videos, they do not perform well with monocular videos. 
NB~\cite{neuralbody} encodes human structure information with latent codes positioned in SMPL vertex's locations, which fails to capture complex human motions.
NHP~\cite{kwon2021neural} does not generalize well in monocular videos due to the information missing from input images.
Thanks to the LBS weight field and deformation field learned in HumanNeRF~\cite{weng_humannerf_2022_cvpr} and AS~\cite{peng2022animatable}, they achieve comparable visualization results as our \nickname{} for 600x more time for training and 472x more time for rendering.
InstantNVR~\cite{geng2023learning} and InstantAvatar~\cite{jiang2023instantavatar} reduce the training time to several minutes with a multi-hashing encoding representation, but they produce slightly worse visualization results than ours and require 45x more rendering time.
DVA~\cite{} improves the rendering speed to 16.5 FPS with efficient mixtures of volumetric primitives.
However, it fails to produce high-quality results and produces obvious artifacts on garments and faces, due to incomplete information within conditional input images and a lack of mechanisms to adaptively control the number of voxels.

\subsection{Ablation Study}
We perform ablation studies with the sequence of 386 from the ZJU\_Mocap data set.
To highlight the effectiveness of each proposed component, we project the 3D human bounding box to each camera plane to obtain the bounding box mask and report these metrics based on the masked area.

\begin{figure*}[h]
    \centering
    \includegraphics[width=18cm]{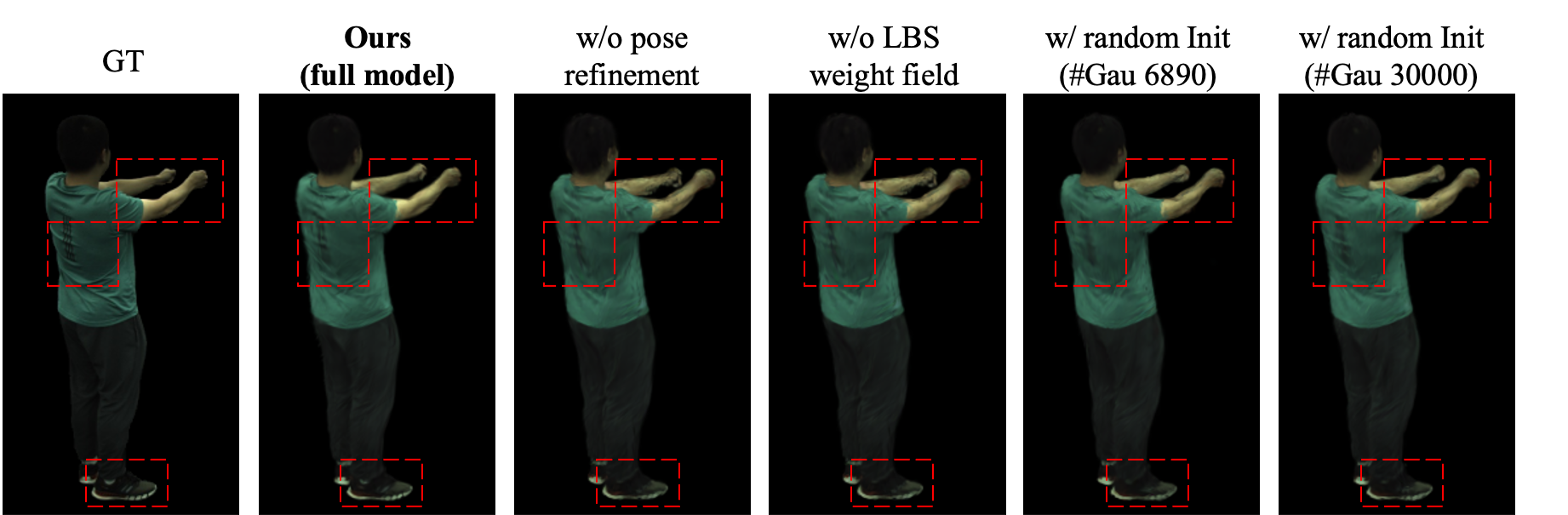}
    \setlength{\abovecaptionskip}{-0.1cm}
    \caption{Qualitative results of ablation study on the 386 sequence of ZJU\_Mocap data set. "w/o pose refinement" and "w/o LBS weight field" mean that we do not adopt the pose refinement or LBS weight field module respectively. "w/ random Init (\#Gau 6890)" and "w/ random Init (\#Gau 30000)" mean that we use random initialization to initialize 6890 or 30000 3D Gaussians respectively.} 
\label{fig: ablation_pose_lbs}
\vspace{-4mm}
\end{figure*}

\noindent \textbf{Pose Refinement and LBS Weight Field.}
Tab.~\ref{tab: ablation_pose_lbs} shows the results of ablating pose refinement and LBS weight field modules.
All models are trained in around 1 minute. 
As shown in Tab.~\ref{tab: ablation_pose_lbs}, pose refinement and LBS weight field modules significantly improve all metrics with negligible computational cost. 
The visualization results in Fig.~\ref{fig: ablation_pose_lbs} show that these two modules can help erase artifacts and learn fine details like cloth wrinkles.
For example, we observe obvious artifacts in hands and fewer cloth wrinkles without using either one of these two modules.

\begin{table}[h]
\setlength{\abovecaptionskip}{0cm}
\caption{Quantitative Results of ablating pose refinement and LBS weight field modules. LPIPS$^*$ = 1000 $\times$ LPIPS.
}
\small
\centering
\label{tab: ablation_pose_lbs}
\setlength{\tabcolsep}{1.0mm}{
\begin{tabular}{lccccc}
\toprule
 & PSNR$\uparrow$ & SSIM$\uparrow$ & LPIPS$^*$$\downarrow$ & Train & FPS\\
\midrule
Ours (full model) & \textbf{28.08} & \textbf{0.886} & \textbf{103.3} & 55s & 189\\
Ours w/o pose refine & 27.94 & 0.884 & 108.8 & 45s & 192\\
Ours w/o LBS field & 27.87 & 0.882 & 108.9 & 52s & 190\\
\bottomrule
\end{tabular}}
\vspace{-3mm}
\end{table}

\noindent \textbf{3D Gaussian Initialization.} 
In Gaussian Splatting, previous research~\cite{kerbl20233d} indicates that initialization plays an important role in the final results. 
For static scenes, there are two commonly adopted initialization approaches, \ie, random initialization or initialization with sparse points from Structure-from-Motion (SfM).  
To incorporate human articulated information into the initialization, we propose to use articulated vertex points from SMPL to initialize our 3D Gaussians in the canonical space.
We compare our proposed articulated initialization with random initialization\footnote{Estimating sparse point cloud from Structure-from-Motion (SfM) requires multi-view inputs, which is infeasible with monocular videos. So we do not compare it with our articulated initialization.} in Tab.~\ref{ablation: 3d_gaussian_init}.
With the same number of 3D Gaussians at initialization, our proposed articulated initialization produces both better quantitative and qualitative results than random initialization as shown in Tab.~\ref{ablation: 3d_gaussian_init} and Fig.~\ref{fig: ablation_pose_lbs}. 
Notably, our articulated initialization converges x10 times much faster than the random initialization, which illustrates the strength of encoding human articulated information into the initialization.
For random initialization, we need to initialize 30k Gaussians to achieve comparable performance as our proposed articulated initialization, while at the cost of 30x times more training time. 

\begin{table}[h]
\vspace{-3mm}
\setlength{\abovecaptionskip}{0cm}
\caption{Quantitative Results of ablating 3D Gaussian initialization. \#Gau denotes the number of 3D Gaussians at initialization.
}
\small
\centering
\label{ablation: 3d_gaussian_init}
\setlength{\tabcolsep}{0.5mm}{
\begin{tabular}{lcccc}
\toprule
 & PSNR$\uparrow$ & SSIM$\uparrow$ & LPIPS$^*$$\downarrow$ & Train\\
\midrule
Articulated Init (w/ \#Gau 6890) & \textbf{28.08} & \textbf{0.886} & \textbf{103.3} & 55s\\
Random Init (w/ \#Gau 6890) & 27.98 & 0.882 & 108.9 & 5m\\
Random Init (w/ \#Gau 30k) & \textbf{28.08} & 0.884 & 104.3 & 30m \\
\bottomrule
\end{tabular}}
\vspace{-3mm}
\end{table}

\noindent \textbf{Split/Clone/Merge/Prune.} 
Adaptive control of 3D Gaussians with split, clone, and prune operations substantially improves the rendering quality, but it trivially leads to a large number of 3D Gaussians (up to 100k), which slows down the whole optimization process. 
As mentioned in Sec.~\ref{sec: efficient_opt}, we propose to use the KL divergence measure to further regulate the split and clone process. 
As shown in Tab.~\ref{ablation: kl_split_clone_merge}, without sacrificing the final performance, we learn a 3D human performer with about 13k 3D Gaussians.
Without sacrificing performance, our KL-based split and clone operations lead to 60x times faster convergence speed when compared with the original split and clone operation proposed in~\cite{kerbl20233d}. 
Furthermore, our novel merge operation merges about 8.5\% of 3D Gaussians without degrading the performance.
Articulated prune operation with SMPL prior introduced in Sec.~\ref{sec: efficient_opt} also helps prune about 0.5k 3D Gaussians far away from the 3D human body. 

\begin{table}[h]
\vspace{-3mm}
\setlength{\abovecaptionskip}{0cm}
\caption{Quantitative Results of ablating our proposed KL-based split/clone operations and a novel merge operation. \#Gau denotes the final number of 3D Gaussians during optimization.
}
\small
\centering
\label{ablation: kl_split_clone_merge}
\setlength{\tabcolsep}{0.8mm}{
\begin{tabular}{lccccc}
\toprule
 & PSNR$\uparrow$ & SSIM$\uparrow$ & LPIPS$^*$$\downarrow$ & \#Gau & Train\\
\midrule
Ours (full model) & \textbf{28.08} & \textbf{0.886} & \textbf{103.3} & \textbf{13162} & \textbf{55s} \\
w/o KL split/clone & 28.01 & 0.885 & 103.1 & 100k & 1h \\
w/o KL merge & \textbf{28.08} & \textbf{0.886} & 103.6 & 14376 & 70s \\
w/o articulated prune & \textbf{28.08} & \textbf{0.886} & 103.4 & 13687 & 57s\\
\bottomrule
\end{tabular}}
\vspace{-3mm}
\end{table}

\section{Discussion and Conclusion}
\label{sec:discussion}

To conclude, we propose \nickname{}, a 3D human model with Gaussian Splatting representation for fast training (1$\sim$2 minutes) and real-time rendering (166 FPS) of 3D humans. 
Specifically, we encode this representation in a canonical space and transform 3D Gaussian from canonical space to posed space with LBS transformation, in which effective pose refinement and LBS weight field modules are also designed to learn fine details of 3D humans.
To enable fast optimization, we initialize and prune 3D Gaussians with 3D human prior, while splitting/cloning 3D Gaussians via Kl divergence guidance, accompanied by a novel merge operation for further speeding up. 
Experiments on monocular videos suggest that our \nickname{} achieves state-of-the-art performance while maintaining both fast training and real-time rendering speed.
In addition, \nickname{} can fast learn 3D human performers with $\sim$13k Gaussians.

\noindent \textbf{Limitations.} 
1). \nickname{} is composed of 3D Gaussians and current pipeline does not support extracting 3D meshes.
How to extract meshes from 3D Gaussians remains a future work to be investigated. 
2). Recovering 3D human details like cloth wrinkles from monocular videos is still a challenging problem to tackle.
One direction is to incorporate physical simulation of garments into current pipelines.

\noindent \textbf{Acknowledgment.} 
We would like to thank Tao Hu from NTU for discussions on the paper framework; Zhaoxi Chen from NTU for discussions on implementing the DVA baseline; Wei Li and Guangcong Wang from NTU for comments on the paper draft; Zhiyuan Yu from ZJU for discussions on the projection matrix used in Gaussian Splatting.

{
    \small
    \bibliographystyle{ieeenat_fullname}
    \bibliography{main}

\begin{thebibliography}{135}
\providecommand{\natexlab}[1]{#1}
\providecommand{\url}[1]{\texttt{#1}}
\expandafter\ifx\csname urlstyle\endcsname\relax
  \providecommand{\doi}[1]{doi: #1}\else
  \providecommand{\doi}{doi: \begingroup \urlstyle{rm}\Url}\fi

\bibitem[AlBahar et~al.(2021)AlBahar, Lu, Yang, Shu, Shechtman, and
  Huang]{albahar2021pose}
Badour AlBahar, Jingwan Lu, Jimei Yang, Zhixin Shu, Eli Shechtman, and Jia-Bin
  Huang.
\newblock Pose with style: Detail-preserving pose-guided image synthesis with
  conditional stylegan.
\newblock \emph{ACM Transactions on Graphics (TOG)}, 40\penalty0 (6):\penalty0
  1--11, 2021.

\bibitem[Aliev et~al.(2020)Aliev, Sevastopolsky, Kolos, Ulyanov, and
  Lempitsky]{aliev2020neural}
Kara-Ali Aliev, Artem Sevastopolsky, Maria Kolos, Dmitry Ulyanov, and Victor
  Lempitsky.
\newblock Neural point-based graphics.
\newblock In \emph{Computer Vision--ECCV 2020: 16th European Conference,
  Glasgow, UK, August 23--28, 2020, Proceedings, Part XXII 16}, pages 696--712.
  Springer, 2020.

\bibitem[Alldieck et~al.(2022)Alldieck, Zanfir, and
  Sminchisescu]{alldieck2022photorealistic}
Thiemo Alldieck, Mihai Zanfir, and Cristian Sminchisescu.
\newblock Photorealistic monocular 3d reconstruction of humans wearing
  clothing.
\newblock In \emph{Proceedings of the IEEE/CVF Conference on Computer Vision
  and Pattern Recognition}, pages 1506--1515, 2022.

\bibitem[Barron et~al.(2022)Barron, Mildenhall, Verbin, Srinivasan, and
  Hedman]{barron2022mip}
Jonathan~T Barron, Ben Mildenhall, Dor Verbin, Pratul~P Srinivasan, and Peter
  Hedman.
\newblock Mip-nerf 360: Unbounded anti-aliased neural radiance fields.
\newblock In \emph{Proceedings of the IEEE/CVF Conference on Computer Vision
  and Pattern Recognition}, pages 5470--5479, 2022.

\bibitem[Bhatnagar et~al.(2020)Bhatnagar, Sminchisescu, Theobalt, and
  Pons-Moll]{bhatnagar2020loopreg}
Bharat~Lal Bhatnagar, Cristian Sminchisescu, Christian Theobalt, and Gerard
  Pons-Moll.
\newblock Loopreg: Self-supervised learning of implicit surface
  correspondences, pose and shape for 3d human mesh registration.
\newblock \emph{Advances in Neural Information Processing Systems},
  33:\penalty0 12909--12922, 2020.

\bibitem[Botsch et~al.(2005)Botsch, Hornung, Zwicker, and
  Kobbelt]{botsch2005high}
Mario Botsch, Alexander Hornung, Matthias Zwicker, and Leif Kobbelt.
\newblock High-quality surface splatting on today's gpus.
\newblock In \emph{Proceedings Eurographics/IEEE VGTC Symposium Point-Based
  Graphics, 2005.}, pages 17--141. IEEE, 2005.

\bibitem[Buehler et~al.(2023)Buehler, Bosse, McMillan, Gortler, and
  Cohen]{buehler2023unstructured}
Chris Buehler, Michael Bosse, Leonard McMillan, Steven Gortler, and Michael
  Cohen.
\newblock Unstructured lumigraph rendering.
\newblock In \emph{Seminal Graphics Papers: Pushing the Boundaries, Volume 2},
  pages 497--504. 2023.

\bibitem[Cha et~al.(2023)Cha, Seo, Ashtari, and Noh]{cha2023generating}
Sihun Cha, Kwanggyoon Seo, Amirsaman Ashtari, and Junyong Noh.
\newblock Generating texture for 3d human avatar from a single image using
  sampling and refinement networks.
\newblock In \emph{Computer Graphics Forum}, pages 385--396. Wiley Online
  Library, 2023.

\bibitem[Chaurasia et~al.(2013)Chaurasia, Duchene, Sorkine-Hornung, and
  Drettakis]{chaurasia2013depth}
Gaurav Chaurasia, Sylvain Duchene, Olga Sorkine-Hornung, and George Drettakis.
\newblock Depth synthesis and local warps for plausible image-based navigation.
\newblock \emph{ACM Transactions on Graphics (TOG)}, 32\penalty0 (3):\penalty0
  1--12, 2013.

\bibitem[Chen et~al.(2021{\natexlab{a}})Chen, Xu, Zhao, Zhang, Xiang, Yu, and
  Su]{chen2021mvsnerf}
Anpei Chen, Zexiang Xu, Fuqiang Zhao, Xiaoshuai Zhang, Fanbo Xiang, Jingyi Yu,
  and Hao Su.
\newblock Mvsnerf: Fast generalizable radiance field reconstruction from
  multi-view stereo.
\newblock \emph{arXiv preprint arXiv:2103.15595}, 2021{\natexlab{a}}.

\bibitem[Chen et~al.(2022)Chen, Xu, Geiger, Yu, and Su]{chen2022tensorf}
Anpei Chen, Zexiang Xu, Andreas Geiger, Jingyi Yu, and Hao Su.
\newblock Tensorf: Tensorial radiance fields.
\newblock In \emph{European Conference on Computer Vision}, pages 333--350.
  Springer, 2022.

\bibitem[Chen et~al.(2021{\natexlab{b}})Chen, Zhang, Kang, Zhe, Bao, Jia, and
  Lu]{chen2021animatable}
Jianchuan Chen, Ying Zhang, Di Kang, Xuefei Zhe, Linchao Bao, Xu Jia, and
  Huchuan Lu.
\newblock Animatable neural radiance fields from monocular rgb videos.
\newblock \emph{arXiv preprint arXiv:2106.13629}, 2021{\natexlab{b}}.

\bibitem[Chen et~al.(2021{\natexlab{c}})Chen, Zheng, Black, Hilliges, and
  Geiger]{chen2021snarf}
Xu Chen, Yufeng Zheng, Michael~J Black, Otmar Hilliges, and Andreas Geiger.
\newblock Snarf: Differentiable forward skinning for animating non-rigid neural
  implicit shapes.
\newblock In \emph{Proceedings of the IEEE/CVF International Conference on
  Computer Vision}, pages 11594--11604, 2021{\natexlab{c}}.

\bibitem[Chen et~al.(2023)Chen, Hong, Mei, Wang, Yang, and
  Liu]{chen2023primdiffusion}
Zhaoxi Chen, Fangzhou Hong, Haiyi Mei, Guangcong Wang, Lei Yang, and Ziwei Liu.
\newblock Primdiffusion: Volumetric primitives diffusion for 3d human
  generation.
\newblock In \emph{Thirty-seventh Conference on Neural Information Processing
  Systems}, 2023.

\bibitem[Cheng et~al.(2023)Cheng, Chen, Fan, Yin, Chen, Cai, Wang, Gao, Yu,
  Lin, et~al.]{cheng2023dna}
Wei Cheng, Ruixiang Chen, Siming Fan, Wanqi Yin, Keyu Chen, Zhongang Cai,
  Jingbo Wang, Yang Gao, Zhengming Yu, Zhengyu Lin, et~al.
\newblock Dna-rendering: A diverse neural actor repository for high-fidelity
  human-centric rendering.
\newblock In \emph{Proceedings of the IEEE/CVF International Conference on
  Computer Vision}, pages 19982--19993, 2023.

\bibitem[Chibane et~al.(2021)Chibane, Bansal, Lazova, and
  Pons-Moll]{chibane2021stereo}
Julian Chibane, Aayush Bansal, Verica Lazova, and Gerard Pons-Moll.
\newblock Stereo radiance fields (srf): Learning view synthesis for sparse
  views of novel scenes.
\newblock In \emph{Proceedings of the IEEE/CVF Conference on Computer Vision
  and Pattern Recognition}, pages 7911--7920, 2021.

\bibitem[Choi et~al.(2022)Choi, Moon, Armando, Leroy, Lee, and
  Rogez]{choi2022mononhr}
Hongsuk Choi, Gyeongsik Moon, Matthieu Armando, Vincent Leroy, Kyoung~Mu Lee,
  and Gr{\'e}gory Rogez.
\newblock Mononhr: Monocular neural human renderer.
\newblock In \emph{2022 International Conference on 3D Vision (3DV)}, pages
  242--251. IEEE, 2022.

\bibitem[Collet et~al.(2015)Collet, Chuang, Sweeney, Gillett, Evseev,
  Calabrese, Hoppe, Kirk, and Sullivan]{collet2015high}
Alvaro Collet, Ming Chuang, Pat Sweeney, Don Gillett, Dennis Evseev, David
  Calabrese, Hugues Hoppe, Adam Kirk, and Steve Sullivan.
\newblock High-quality streamable free-viewpoint video.
\newblock \emph{ACM Transactions on Graphics (ToG)}, 34\penalty0 (4):\penalty0
  1--13, 2015.

\bibitem[Darmon et~al.(2022)Darmon, Bascle, Devaux, Monasse, and
  Aubry]{darmon2022improving}
Fran{\c{c}}ois Darmon, B{\'e}n{\'e}dicte Bascle, Jean-Cl{\'e}ment Devaux,
  Pascal Monasse, and Mathieu Aubry.
\newblock Improving neural implicit surfaces geometry with patch warping.
\newblock In \emph{Proceedings of the IEEE/CVF Conference on Computer Vision
  and Pattern Recognition}, pages 6260--6269, 2022.

\bibitem[Deng et~al.(2020)Deng, Lewis, Jeruzalski, Pons-Moll, Hinton, Norouzi,
  and Tagliasacchi]{deng2020nasa}
Boyang Deng, John~P Lewis, Timothy Jeruzalski, Gerard Pons-Moll, Geoffrey
  Hinton, Mohammad Norouzi, and Andrea Tagliasacchi.
\newblock Nasa neural articulated shape approximation.
\newblock In \emph{Computer Vision--ECCV 2020: 16th European Conference,
  Glasgow, UK, August 23--28, 2020, Proceedings, Part VII 16}, pages 612--628.
  Springer, 2020.

\bibitem[Deng et~al.(2021)Deng, Liu, Zhu, and Ramanan]{deng2021depth}
Kangle Deng, Andrew Liu, Jun-Yan Zhu, and Deva Ramanan.
\newblock Depth-supervised nerf: Fewer views and faster training for free.
\newblock \emph{arXiv preprint arXiv:2107.02791}, 2021.

\bibitem[Dou et~al.(2016)Dou, Khamis, Degtyarev, Davidson, Fanello, Kowdle,
  Escolano, Rhemann, Kim, Taylor, et~al.]{dou2016fusion4d}
Mingsong Dou, Sameh Khamis, Yury Degtyarev, Philip Davidson, Sean~Ryan Fanello,
  Adarsh Kowdle, Sergio~Orts Escolano, Christoph Rhemann, David Kim, Jonathan
  Taylor, et~al.
\newblock Fusion4d: Real-time performance capture of challenging scenes.
\newblock \emph{ACM Transactions on Graphics (ToG)}, 35\penalty0 (4):\penalty0
  1--13, 2016.

\bibitem[Eisemann et~al.(2008)Eisemann, De~Decker, Magnor, Bekaert, De~Aguiar,
  Ahmed, Theobalt, and Sellent]{eisemann2008floating}
Martin Eisemann, Bert De~Decker, Marcus Magnor, Philippe Bekaert, Edilson
  De~Aguiar, Naveed Ahmed, Christian Theobalt, and Anita Sellent.
\newblock Floating textures.
\newblock In \emph{Computer graphics forum}, pages 409--418. Wiley Online
  Library, 2008.

\bibitem[Flynn et~al.(2016)Flynn, Neulander, Philbin, and
  Snavely]{flynn2016deepstereo}
John Flynn, Ivan Neulander, James Philbin, and Noah Snavely.
\newblock Deepstereo: Learning to predict new views from the world's imagery.
\newblock In \emph{Proceedings of the IEEE conference on computer vision and
  pattern recognition}, pages 5515--5524, 2016.

\bibitem[Fridovich-Keil et~al.(2022)Fridovich-Keil, Yu, Tancik, Chen, Recht,
  and Kanazawa]{fridovich2022plenoxels}
Sara Fridovich-Keil, Alex Yu, Matthew Tancik, Qinhong Chen, Benjamin Recht, and
  Angjoo Kanazawa.
\newblock Plenoxels: Radiance fields without neural networks.
\newblock In \emph{Proceedings of the IEEE/CVF Conference on Computer Vision
  and Pattern Recognition}, pages 5501--5510, 2022.

\bibitem[Fu et~al.(2022)Fu, Li, Jiang, Lin, Qian, Loy, Wu, and
  Liu]{fu2022stylegan}
Jianglin Fu, Shikai Li, Yuming Jiang, Kwan-Yee Lin, Chen Qian, Chen~Change Loy,
  Wayne Wu, and Ziwei Liu.
\newblock Stylegan-human: A data-centric odyssey of human generation.
\newblock In \emph{European Conference on Computer Vision}, pages 1--19.
  Springer, 2022.

\bibitem[Fu et~al.(2023)Fu, Li, Jiang, Lin, Wu, and Liu]{fu2023unitedhuman}
Jianglin Fu, Shikai Li, Yuming Jiang, Kwan-Yee Lin, Wayne Wu, and Ziwei Liu.
\newblock Unitedhuman: Harnessing multi-source data for high-resolution human
  generation.
\newblock In \emph{Proceedings of the IEEE/CVF International Conference on
  Computer Vision}, pages 7301--7311, 2023.

\bibitem[Gao et~al.(2022)Gao, Yang, Kim, Peng, Liu, and Tong]{gao2022mps}
Xiangjun Gao, Jiaolong Yang, Jongyoo Kim, Sida Peng, Zicheng Liu, and Xin Tong.
\newblock Mps-nerf: Generalizable 3d human rendering from multiview images.
\newblock \emph{IEEE Transactions on Pattern Analysis and Machine
  Intelligence}, 2022.

\bibitem[Garbin et~al.(2021)Garbin, Kowalski, Johnson, Shotton, and
  Valentin]{garbin2021fastnerf}
Stephan~J Garbin, Marek Kowalski, Matthew Johnson, Jamie Shotton, and Julien
  Valentin.
\newblock Fastnerf: High-fidelity neural rendering at 200fps.
\newblock In \emph{Proceedings of the IEEE/CVF International Conference on
  Computer Vision}, pages 14346--14355, 2021.

\bibitem[Geng et~al.(2023)Geng, Peng, Xu, Bao, and Zhou]{geng2023learning}
Chen Geng, Sida Peng, Zhen Xu, Hujun Bao, and Xiaowei Zhou.
\newblock Learning neural volumetric representations of dynamic humans in
  minutes.
\newblock In \emph{Proceedings of the IEEE/CVF Conference on Computer Vision
  and Pattern Recognition}, pages 8759--8770, 2023.

\bibitem[Goesele et~al.(2007)Goesele, Snavely, Curless, Hoppe, and
  Seitz]{goesele2007multi}
Michael Goesele, Noah Snavely, Brian Curless, Hugues Hoppe, and Steven~M Seitz.
\newblock Multi-view stereo for community photo collections.
\newblock In \emph{2007 IEEE 11th International Conference on Computer Vision},
  pages 1--8. IEEE, 2007.

\bibitem[Gortler et~al.(2023)Gortler, Grzeszczuk, Szeliski, and
  Cohen]{gortler2023lumigraph}
Steven~J Gortler, Radek Grzeszczuk, Richard Szeliski, and Michael~F Cohen.
\newblock The lumigraph.
\newblock In \emph{Seminal Graphics Papers: Pushing the Boundaries, Volume 2},
  pages 453--464. 2023.

\bibitem[Gross and Pfister(2011)]{gross2011point}
Markus Gross and Hanspeter Pfister.
\newblock \emph{Point-based graphics}.
\newblock Elsevier, 2011.

\bibitem[Grossman and Dally(1998)]{grossman1998point}
Jeffrey~P Grossman and William~J Dally.
\newblock Point sample rendering.
\newblock In \emph{Rendering Techniques’ 98: Proceedings of the Eurographics
  Workshop in Vienna, Austria, June 29—July 1, 1998 9}, pages 181--192.
  Springer, 1998.

\bibitem[Guo et~al.(2019)Guo, Lincoln, Davidson, Busch, Yu, Whalen, Harvey,
  Orts-Escolano, Pandey, Dourgarian, et~al.]{guo2019relightables}
Kaiwen Guo, Peter Lincoln, Philip Davidson, Jay Busch, Xueming Yu, Matt Whalen,
  Geoff Harvey, Sergio Orts-Escolano, Rohit Pandey, Jason Dourgarian, et~al.
\newblock The relightables: Volumetric performance capture of humans with
  realistic relighting.
\newblock \emph{ACM Transactions on Graphics (ToG)}, 38\penalty0 (6):\penalty0
  1--19, 2019.

\bibitem[Habermann et~al.(2020)Habermann, Xu, Zollhofer, Pons-Moll, and
  Theobalt]{habermann2020deepcap}
Marc Habermann, Weipeng Xu, Michael Zollhofer, Gerard Pons-Moll, and Christian
  Theobalt.
\newblock Deepcap: Monocular human performance capture using weak supervision.
\newblock In \emph{Proceedings of the IEEE/CVF Conference on Computer Vision
  and Pattern Recognition}, pages 5052--5063, 2020.

\bibitem[Habermann et~al.(2021)Habermann, Liu, Xu, Zollhoefer, Pons-Moll, and
  Theobalt]{habermann2021real}
Marc Habermann, Lingjie Liu, Weipeng Xu, Michael Zollhoefer, Gerard Pons-Moll,
  and Christian Theobalt.
\newblock Real-time deep dynamic characters.
\newblock \emph{ACM Transactions on Graphics (ToG)}, 40\penalty0 (4):\penalty0
  1--16, 2021.

\bibitem[Hedman et~al.(2018)Hedman, Philip, Price, Frahm, Drettakis, and
  Brostow]{hedman2018deep}
Peter Hedman, Julien Philip, True Price, Jan-Michael Frahm, George Drettakis,
  and Gabriel Brostow.
\newblock Deep blending for free-viewpoint image-based rendering.
\newblock \emph{ACM Transactions on Graphics (ToG)}, 37\penalty0 (6):\penalty0
  1--15, 2018.

\bibitem[Henzler et~al.(2019)Henzler, Mitra, and Ritschel]{henzler2019escaping}
Philipp Henzler, Niloy~J Mitra, and Tobias Ritschel.
\newblock Escaping plato's cave: 3d shape from adversarial rendering.
\newblock In \emph{Proceedings of the IEEE/CVF International Conference on
  Computer Vision}, pages 9984--9993, 2019.

\bibitem[Hu et~al.(2023)Hu, Hong, Pan, Mei, Yang, and Liu]{hu2023sherf}
Shoukang Hu, Fangzhou Hong, Liang Pan, Haiyi Mei, Lei Yang, and Ziwei Liu.
\newblock Sherf: Generalizable human nerf from a single image.
\newblock \emph{arXiv preprint arXiv:2303.12791}, 2023.

\bibitem[Huang et~al.(2023)Huang, Yi, Xiu, Liao, Tang, Cai, and
  Thies]{huang2023tech}
Yangyi Huang, Hongwei Yi, Yuliang Xiu, Tingting Liao, Jiaxiang Tang, Deng Cai,
  and Justus Thies.
\newblock Tech: Text-guided reconstruction of lifelike clothed humans.
\newblock \emph{arXiv preprint arXiv:2308.08545}, 2023.

\bibitem[Huang et~al.(2020)Huang, Xu, Lassner, Li, and Tung]{huang2020arch}
Zeng Huang, Yuanlu Xu, Christoph Lassner, Hao Li, and Tony Tung.
\newblock Arch: Animatable reconstruction of clothed humans.
\newblock In \emph{Proceedings of the IEEE/CVF Conference on Computer Vision
  and Pattern Recognition}, pages 3093--3102, 2020.

\bibitem[Jain et~al.(2021)Jain, Tancik, and Abbeel]{jain2021putting}
Ajay Jain, Matthew Tancik, and Pieter Abbeel.
\newblock Putting nerf on a diet: Semantically consistent few-shot view
  synthesis.
\newblock In \emph{Proceedings of the IEEE/CVF International Conference on
  Computer Vision}, pages 5885--5894, 2021.

\bibitem[Jang and Agapito(2021)]{jang2021codenerf}
Wonbong Jang and Lourdes Agapito.
\newblock Codenerf: Disentangled neural radiance fields for object categories.
\newblock In \emph{Proceedings of the IEEE/CVF International Conference on
  Computer Vision}, pages 12949--12958, 2021.

\bibitem[Jiang et~al.(2022{\natexlab{a}})Jiang, Hong, Bao, and
  Zhang]{jiang2022selfrecon}
Boyi Jiang, Yang Hong, Hujun Bao, and Juyong Zhang.
\newblock Selfrecon: Self reconstruction your digital avatar from monocular
  video.
\newblock In \emph{Proceedings of the IEEE/CVF Conference on Computer Vision
  and Pattern Recognition}, pages 5605--5615, 2022{\natexlab{a}}.

\bibitem[Jiang et~al.(2023{\natexlab{a}})Jiang, Chen, Song, and
  Hilliges]{jiang2023instantavatar}
Tianjian Jiang, Xu Chen, Jie Song, and Otmar Hilliges.
\newblock Instantavatar: Learning avatars from monocular video in 60 seconds.
\newblock In \emph{Proceedings of the IEEE/CVF Conference on Computer Vision
  and Pattern Recognition}, pages 16922--16932, 2023{\natexlab{a}}.

\bibitem[Jiang et~al.(2022{\natexlab{b}})Jiang, Yi, Samei, Tuzel, and
  Ranjan]{jiang2022neuman}
Wei Jiang, Kwang~Moo Yi, Golnoosh Samei, Oncel Tuzel, and Anurag Ranjan.
\newblock Neuman: Neural human radiance field from a single video.
\newblock \emph{arXiv preprint arXiv:2203.12575}, 2022{\natexlab{b}}.

\bibitem[Jiang et~al.(2022{\natexlab{c}})Jiang, Yang, Qiu, Wu, Loy, and
  Liu]{jiang2022text2human}
Yuming Jiang, Shuai Yang, Haonan Qiu, Wayne Wu, Chen~Change Loy, and Ziwei Liu.
\newblock Text2human: Text-driven controllable human image generation.
\newblock \emph{ACM Transactions on Graphics (TOG)}, 41\penalty0 (4):\penalty0
  1--11, 2022{\natexlab{c}}.

\bibitem[Jiang et~al.(2023{\natexlab{b}})Jiang, Yang, Koh, Wu, Loy, and
  Liu]{jiang2023text2performer}
Yuming Jiang, Shuai Yang, Tong~Liang Koh, Wayne Wu, Chen~Change Loy, and Ziwei
  Liu.
\newblock Text2performer: Text-driven human video generation.
\newblock \emph{arXiv preprint arXiv:2304.08483}, 2023{\natexlab{b}}.

\bibitem[Johari et~al.(2022)Johari, Lepoittevin, and
  Fleuret]{johari2022geonerf}
Mohammad~Mahdi Johari, Yann Lepoittevin, and Fran{\c{c}}ois Fleuret.
\newblock Geonerf: Generalizing nerf with geometry priors.
\newblock In \emph{Proceedings of the IEEE/CVF Conference on Computer Vision
  and Pattern Recognition}, pages 18365--18375, 2022.

\bibitem[Kajiya and Von~Herzen(1984)]{kajiya1984ray}
James~T Kajiya and Brian~P Von~Herzen.
\newblock Ray tracing volume densities.
\newblock \emph{ACM SIGGRAPH computer graphics}, 18\penalty0 (3):\penalty0
  165--174, 1984.

\bibitem[Kerbl et~al.(2023)Kerbl, Kopanas, Leimk{\"u}hler, and
  Drettakis]{kerbl20233d}
Bernhard Kerbl, Georgios Kopanas, Thomas Leimk{\"u}hler, and George Drettakis.
\newblock 3d gaussian splatting for real-time radiance field rendering.
\newblock \emph{ACM Transactions on Graphics (ToG)}, 42\penalty0 (4):\penalty0
  1--14, 2023.

\bibitem[Kopanas et~al.(2021)Kopanas, Philip, Leimk{\"u}hler, and
  Drettakis]{kopanas2021point}
Georgios Kopanas, Julien Philip, Thomas Leimk{\"u}hler, and George Drettakis.
\newblock Point-based neural rendering with per-view optimization.
\newblock In \emph{Computer Graphics Forum}, pages 29--43. Wiley Online
  Library, 2021.

\bibitem[Kopanas et~al.(2022)Kopanas, Leimk{\"u}hler, Rainer, Jambon, and
  Drettakis]{kopanas2022neural}
Georgios Kopanas, Thomas Leimk{\"u}hler, Gilles Rainer, Cl{\'e}ment Jambon, and
  George Drettakis.
\newblock Neural point catacaustics for novel-view synthesis of reflections.
\newblock \emph{ACM Transactions on Graphics (TOG)}, 41\penalty0 (6):\penalty0
  1--15, 2022.

\bibitem[Kwon et~al.(2021)Kwon, Kim, Ceylan, and Fuchs]{kwon2021neural}
Youngjoong Kwon, Dahun Kim, Duygu Ceylan, and Henry Fuchs.
\newblock Neural human performer: Learning generalizable radiance fields for
  human performance rendering.
\newblock \emph{Advances in Neural Information Processing Systems},
  34:\penalty0 24741--24752, 2021.

\bibitem[Lassner and Zollhofer(2021)]{lassner2021pulsar}
Christoph Lassner and Michael Zollhofer.
\newblock Pulsar: Efficient sphere-based neural rendering.
\newblock In \emph{Proceedings of the IEEE/CVF Conference on Computer Vision
  and Pattern Recognition}, pages 1440--1449, 2021.

\bibitem[Lei et~al.(2023)Lei, Wang, Pavlakos, Liu, and Daniilidis]{lei2023gart}
Jiahui Lei, Yufu Wang, Georgios Pavlakos, Lingjie Liu, and Kostas Daniilidis.
\newblock Gart: Gaussian articulated template models.
\newblock \emph{arXiv preprint arXiv:2311.16099}, 2023.

\bibitem[Levoy and Hanrahan(2023)]{levoy2023light}
Marc Levoy and Pat Hanrahan.
\newblock Light field rendering.
\newblock In \emph{Seminal Graphics Papers: Pushing the Boundaries, Volume 2},
  pages 441--452. 2023.

\bibitem[Lewis et~al.(2021)Lewis, Varadharajan, and
  Kemelmacher-Shlizerman]{lewis2021tryongan}
Kathleen~M Lewis, Srivatsan Varadharajan, and Ira Kemelmacher-Shlizerman.
\newblock Tryongan: Body-aware try-on via layered interpolation.
\newblock \emph{ACM Transactions on Graphics (TOG)}, 40\penalty0 (4):\penalty0
  1--10, 2021.

\bibitem[Li et~al.(2022)Li, Tancik, and Kanazawa]{li2022nerfacc}
Ruilong Li, Matthew Tancik, and Angjoo Kanazawa.
\newblock Nerfacc: A general nerf acceleration toolbox.
\newblock \emph{arXiv preprint arXiv:2210.04847}, 2022.

\bibitem[Liao et~al.(2023)Liao, Zhang, Xiu, Yi, Liu, Qi, Zhang, Wang, Zhu, and
  Lei]{liao2023high}
Tingting Liao, Xiaomei Zhang, Yuliang Xiu, Hongwei Yi, Xudong Liu, Guo-Jun Qi,
  Yong Zhang, Xuan Wang, Xiangyu Zhu, and Zhen Lei.
\newblock High-fidelity clothed avatar reconstruction from a single image.
\newblock In \emph{Proceedings of the IEEE/CVF Conference on Computer Vision
  and Pattern Recognition}, pages 8662--8672, 2023.

\bibitem[Lin et~al.(2022)Lin, Peng, Xu, Yan, Shuai, Bao, and
  Zhou]{lin2022efficient}
Haotong Lin, Sida Peng, Zhen Xu, Yunzhi Yan, Qing Shuai, Hujun Bao, and Xiaowei
  Zhou.
\newblock Efficient neural radiance fields for interactive free-viewpoint
  video.
\newblock In \emph{SIGGRAPH Asia 2022 Conference Papers}, pages 1--9, 2022.

\bibitem[Liu et~al.(2020)Liu, Gu, Zaw~Lin, Chua, and Theobalt]{liu2020neural}
Lingjie Liu, Jiatao Gu, Kyaw Zaw~Lin, Tat-Seng Chua, and Christian Theobalt.
\newblock Neural sparse voxel fields.
\newblock \emph{Advances in Neural Information Processing Systems},
  33:\penalty0 15651--15663, 2020.

\bibitem[Liu et~al.(2021)Liu, Habermann, Rudnev, Sarkar, Gu, and
  Theobalt]{liu2021neural}
Lingjie Liu, Marc Habermann, Viktor Rudnev, Kripasindhu Sarkar, Jiatao Gu, and
  Christian Theobalt.
\newblock Neural actor: Neural free-view synthesis of human actors with pose
  control.
\newblock \emph{ACM transactions on graphics (TOG)}, 40\penalty0 (6):\penalty0
  1--16, 2021.

\bibitem[Liu et~al.(2022)Liu, Peng, Liu, Wang, Wang, Theobalt, Zhou, and
  Wang]{liu2022neural}
Yuan Liu, Sida Peng, Lingjie Liu, Qianqian Wang, Peng Wang, Christian Theobalt,
  Xiaowei Zhou, and Wenping Wang.
\newblock Neural rays for occlusion-aware image-based rendering.
\newblock In \emph{Proceedings of the IEEE/CVF Conference on Computer Vision
  and Pattern Recognition}, pages 7824--7833, 2022.

\bibitem[Lombardi et~al.(2021)Lombardi, Simon, Schwartz, Zollhoefer, Sheikh,
  and Saragih]{lombardi2021mixture}
Stephen Lombardi, Tomas Simon, Gabriel Schwartz, Michael Zollhoefer, Yaser
  Sheikh, and Jason Saragih.
\newblock Mixture of volumetric primitives for efficient neural rendering.
\newblock \emph{ACM Transactions on Graphics (ToG)}, 40\penalty0 (4):\penalty0
  1--13, 2021.

\bibitem[Loper et~al.(2015)Loper, Mahmood, Romero, Pons-Moll, and
  Black]{SMPL:2015}
Matthew Loper, Naureen Mahmood, Javier Romero, Gerard Pons-Moll, and Michael~J.
  Black.
\newblock {SMPL}: A skinned multi-person linear model.
\newblock \emph{ACM Trans. Graphics (Proc. SIGGRAPH Asia)}, 34\penalty0
  (6):\penalty0 248:1--248:16, 2015.

\bibitem[Luiten et~al.(2023)Luiten, Kopanas, Leibe, and
  Ramanan]{luiten2023dynamic}
Jonathon Luiten, Georgios Kopanas, Bastian Leibe, and Deva Ramanan.
\newblock Dynamic 3d gaussians: Tracking by persistent dynamic view synthesis.
\newblock \emph{arXiv preprint arXiv:2308.09713}, 2023.

\bibitem[Max(1995)]{max1995optical}
Nelson Max.
\newblock Optical models for direct volume rendering.
\newblock \emph{IEEE Transactions on Visualization and Computer Graphics},
  1\penalty0 (2):\penalty0 99--108, 1995.

\bibitem[Men et~al.(2020)Men, Mao, Jiang, Ma, and Lian]{men2020controllable}
Yifang Men, Yiming Mao, Yuning Jiang, Wei-Ying Ma, and Zhouhui Lian.
\newblock Controllable person image synthesis with attribute-decomposed gan.
\newblock In \emph{Proceedings of the IEEE/CVF conference on computer vision
  and pattern recognition}, pages 5084--5093, 2020.

\bibitem[Mihajlovic et~al.(2021)Mihajlovic, Zhang, Black, and
  Tang]{mihajlovic2021leap}
Marko Mihajlovic, Yan Zhang, Michael~J Black, and Siyu Tang.
\newblock Leap: Learning articulated occupancy of people.
\newblock In \emph{Proceedings of the IEEE/CVF Conference on Computer Vision
  and Pattern Recognition}, pages 10461--10471, 2021.

\bibitem[Mildenhall et~al.(2021)Mildenhall, Srinivasan, Tancik, Barron,
  Ramamoorthi, and Ng]{mildenhall2021nerf}
Ben Mildenhall, Pratul~P Srinivasan, Matthew Tancik, Jonathan~T Barron, Ravi
  Ramamoorthi, and Ren Ng.
\newblock Nerf: Representing scenes as neural radiance fields for view
  synthesis.
\newblock \emph{Communications of the ACM}, 65\penalty0 (1):\penalty0 99--106,
  2021.

\bibitem[M{\"u}ller et~al.(2022)M{\"u}ller, Evans, Schied, and
  Keller]{muller2022instant}
Thomas M{\"u}ller, Alex Evans, Christoph Schied, and Alexander Keller.
\newblock Instant neural graphics primitives with a multiresolution hash
  encoding.
\newblock \emph{ACM Transactions on Graphics (ToG)}, 41\penalty0 (4):\penalty0
  1--15, 2022.

\bibitem[Niemeyer et~al.(2021)Niemeyer, Barron, Mildenhall, Sajjadi, Geiger,
  and Radwan]{niemeyer2021regnerf}
Michael Niemeyer, Jonathan~T Barron, Ben Mildenhall, Mehdi~SM Sajjadi, Andreas
  Geiger, and Noha Radwan.
\newblock Regnerf: Regularizing neural radiance fields for view synthesis from
  sparse inputs.
\newblock \emph{arXiv preprint arXiv:2112.00724}, 2021.

\bibitem[Noguchi et~al.(2021)Noguchi, Sun, Lin, and Harada]{noguchi2021neural}
Atsuhiro Noguchi, Xiao Sun, Stephen Lin, and Tatsuya Harada.
\newblock Neural articulated radiance field.
\newblock In \emph{Proceedings of the IEEE/CVF International Conference on
  Computer Vision}, pages 5762--5772, 2021.

\bibitem[Paszke et~al.(2019)Paszke, Gross, Massa, Lerer, Bradbury, Chanan,
  Killeen, Lin, Gimelshein, Antiga, et~al.]{paszke2019pytorch}
Adam Paszke, Sam Gross, Francisco Massa, Adam Lerer, James Bradbury, Gregory
  Chanan, Trevor Killeen, Zeming Lin, Natalia Gimelshein, Luca Antiga, et~al.
\newblock Pytorch: An imperative style, high-performance deep learning library.
\newblock \emph{Advances in neural information processing systems}, 32, 2019.

\bibitem[Peng et~al.(2021{\natexlab{a}})Peng, Dong, Wang, Zhang, Shuai, Zhou,
  and Bao]{peng2021animatable}
Sida Peng, Junting Dong, Qianqian Wang, Shangzhan Zhang, Qing Shuai, Xiaowei
  Zhou, and Hujun Bao.
\newblock Animatable neural radiance fields for modeling dynamic human bodies.
\newblock In \emph{ICCV}, 2021{\natexlab{a}}.

\bibitem[Peng et~al.(2021{\natexlab{b}})Peng, Zhang, Xu, Wang, Shuai, Bao, and
  Zhou]{neuralbody}
Sida Peng, Yuanqing Zhang, Yinghao Xu, Qianqian Wang, Qing Shuai, Hujun Bao,
  and Xiaowei Zhou.
\newblock Neural body: Implicit neural representations with structured latent
  codes for novel view synthesis of dynamic humans.
\newblock In \emph{Proceedings of the IEEE/CVF Conference on Computer Vision
  and Pattern Recognition}, pages 9054--9063, 2021{\natexlab{b}}.

\bibitem[Peng et~al.(2022)Peng, Zhang, Xu, Geng, Jiang, Bao, and
  Zhou]{peng2022animatable}
Sida Peng, Shangzhan Zhang, Zhen Xu, Chen Geng, Boyi Jiang, Hujun Bao, and
  Xiaowei Zhou.
\newblock Animatable neural implict surfaces for creating avatars from videos.
\newblock \emph{arXiv preprint arXiv:2203.08133}, 4\penalty0 (5), 2022.

\bibitem[Penner and Zhang(2017)]{penner2017soft}
Eric Penner and Li Zhang.
\newblock Soft 3d reconstruction for view synthesis.
\newblock \emph{ACM Transactions on Graphics (TOG)}, 36\penalty0 (6):\penalty0
  1--11, 2017.

\bibitem[Pfister et~al.(2000)Pfister, Zwicker, Van~Baar, and
  Gross]{pfister2000surfels}
Hanspeter Pfister, Matthias Zwicker, Jeroen Van~Baar, and Markus Gross.
\newblock Surfels: Surface elements as rendering primitives.
\newblock In \emph{Proceedings of the 27th annual conference on Computer
  graphics and interactive techniques}, pages 335--342, 2000.

\bibitem[Reiser et~al.(2021)Reiser, Peng, Liao, and Geiger]{reiser2021kilonerf}
Christian Reiser, Songyou Peng, Yiyi Liao, and Andreas Geiger.
\newblock Kilonerf: Speeding up neural radiance fields with thousands of tiny
  mlps.
\newblock In \emph{Proceedings of the IEEE/CVF International Conference on
  Computer Vision}, pages 14335--14345, 2021.

\bibitem[Rematas et~al.(2021)Rematas, Martin-Brualla, and
  Ferrari]{rematas2021sharf}
Konstantinos Rematas, Ricardo Martin-Brualla, and Vittorio Ferrari.
\newblock Sharf: Shape-conditioned radiance fields from a single view.
\newblock \emph{arXiv preprint arXiv:2102.08860}, 2021.

\bibitem[Remelli et~al.(2022)Remelli, Bagautdinov, Saito, Wu, Simon, Wei, Guo,
  Cao, Prada, Saragih, et~al.]{remelli2022drivable}
Edoardo Remelli, Timur Bagautdinov, Shunsuke Saito, Chenglei Wu, Tomas Simon,
  Shih-En Wei, Kaiwen Guo, Zhe Cao, Fabian Prada, Jason Saragih, et~al.
\newblock Drivable volumetric avatars using texel-aligned features.
\newblock In \emph{ACM SIGGRAPH 2022 Conference Proceedings}, pages 1--9, 2022.

\bibitem[Ren et~al.(2002)Ren, Pfister, and Zwicker]{ren2002object}
Liu Ren, Hanspeter Pfister, and Matthias Zwicker.
\newblock Object space ewa surface splatting: A hardware accelerated approach
  to high quality point rendering.
\newblock In \emph{Computer Graphics Forum}, pages 461--470. Wiley Online
  Library, 2002.

\bibitem[Rhodin et~al.(2015)Rhodin, Robertini, Richardt, Seidel, and
  Theobalt]{rhodin2015versatile}
Helge Rhodin, Nadia Robertini, Christian Richardt, Hans-Peter Seidel, and
  Christian Theobalt.
\newblock A versatile scene model with differentiable visibility applied to
  generative pose estimation.
\newblock In \emph{Proceedings of the IEEE International Conference on Computer
  Vision}, pages 765--773, 2015.

\bibitem[Riegler and Koltun(2020)]{riegler2020free}
Gernot Riegler and Vladlen Koltun.
\newblock Free view synthesis.
\newblock In \emph{Computer Vision--ECCV 2020: 16th European Conference,
  Glasgow, UK, August 23--28, 2020, Proceedings, Part XIX 16}, pages 623--640.
  Springer, 2020.

\bibitem[R{\"u}ckert et~al.(2022)R{\"u}ckert, Franke, and
  Stamminger]{ruckert2022adop}
Darius R{\"u}ckert, Linus Franke, and Marc Stamminger.
\newblock Adop: Approximate differentiable one-pixel point rendering.
\newblock \emph{ACM Transactions on Graphics (ToG)}, 41\penalty0 (4):\penalty0
  1--14, 2022.

\bibitem[Sainz and Pajarola(2004)]{sainz2004point}
Miguel Sainz and Renato Pajarola.
\newblock Point-based rendering techniques.
\newblock \emph{Computers \& Graphics}, 28\penalty0 (6):\penalty0 869--879,
  2004.

\bibitem[Saito et~al.(2019)Saito, Huang, Natsume, Morishima, Kanazawa, and
  Li]{saito2019pifu}
Shunsuke Saito, Zeng Huang, Ryota Natsume, Shigeo Morishima, Angjoo Kanazawa,
  and Hao Li.
\newblock Pifu: Pixel-aligned implicit function for high-resolution clothed
  human digitization.
\newblock In \emph{Proceedings of the IEEE/CVF international conference on
  computer vision}, pages 2304--2314, 2019.

\bibitem[Saito et~al.(2020)Saito, Simon, Saragih, and Joo]{saito2020pifuhd}
Shunsuke Saito, Tomas Simon, Jason Saragih, and Hanbyul Joo.
\newblock Pifuhd: Multi-level pixel-aligned implicit function for
  high-resolution 3d human digitization.
\newblock In \emph{Proceedings of the IEEE/CVF Conference on Computer Vision
  and Pattern Recognition}, pages 84--93, 2020.

\bibitem[Saito et~al.(2021)Saito, Yang, Ma, and Black]{saito2021scanimate}
Shunsuke Saito, Jinlong Yang, Qianli Ma, and Michael~J Black.
\newblock Scanimate: Weakly supervised learning of skinned clothed avatar
  networks.
\newblock In \emph{Proceedings of the IEEE/CVF Conference on Computer Vision
  and Pattern Recognition}, pages 2886--2897, 2021.

\bibitem[Sara et~al.(2019)Sara, Akter, and Uddin]{sara2019image}
Umme Sara, Morium Akter, and Mohammad~Shorif Uddin.
\newblock Image quality assessment through fsim, ssim, mse and psnr—a
  comparative study.
\newblock \emph{Journal of Computer and Communications}, 7\penalty0
  (3):\penalty0 8--18, 2019.

\bibitem[Sarkar et~al.(2021)Sarkar, Liu, Golyanik, and
  Theobalt]{sarkar2021humangan}
Kripasindhu Sarkar, Lingjie Liu, Vladislav Golyanik, and Christian Theobalt.
\newblock Humangan: A generative model of human images.
\newblock In \emph{2021 International Conference on 3D Vision (3DV)}, pages
  258--267. IEEE, 2021.

\bibitem[Schonberger and Frahm(2016)]{schonberger2016structure}
Johannes~L Schonberger and Jan-Michael Frahm.
\newblock Structure-from-motion revisited.
\newblock In \emph{Proceedings of the IEEE conference on computer vision and
  pattern recognition}, pages 4104--4113, 2016.

\bibitem[Sch{\"o}nberger et~al.(2016)Sch{\"o}nberger, Zheng, Frahm, and
  Pollefeys]{schonberger2016pixelwise}
Johannes~L Sch{\"o}nberger, Enliang Zheng, Jan-Michael Frahm, and Marc
  Pollefeys.
\newblock Pixelwise view selection for unstructured multi-view stereo.
\newblock In \emph{Computer Vision--ECCV 2016: 14th European Conference,
  Amsterdam, The Netherlands, October 11-14, 2016, Proceedings, Part III 14},
  pages 501--518. Springer, 2016.

\bibitem[Sitzmann et~al.(2019)Sitzmann, Thies, Heide, Nie{\ss}ner, Wetzstein,
  and Zollhofer]{sitzmann2019deepvoxels}
Vincent Sitzmann, Justus Thies, Felix Heide, Matthias Nie{\ss}ner, Gordon
  Wetzstein, and Michael Zollhofer.
\newblock Deepvoxels: Learning persistent 3d feature embeddings.
\newblock In \emph{Proceedings of the IEEE/CVF Conference on Computer Vision
  and Pattern Recognition}, pages 2437--2446, 2019.

\bibitem[Snavely et~al.(2006)Snavely, Seitz, and Szeliski]{snavely2006photo}
Noah Snavely, Steven~M Seitz, and Richard Szeliski.
\newblock Photo tourism: exploring photo collections in 3d.
\newblock In \emph{ACM siggraph 2006 papers}, pages 835--846. 2006.

\bibitem[Stoll et~al.(2011)Stoll, Hasler, Gall, Seidel, and
  Theobalt]{stoll2011fast}
Carsten Stoll, Nils Hasler, Juergen Gall, Hans-Peter Seidel, and Christian
  Theobalt.
\newblock Fast articulated motion tracking using a sums of gaussians body
  model.
\newblock In \emph{2011 International Conference on Computer Vision}, pages
  951--958. IEEE, 2011.

\bibitem[Su et~al.(2021)Su, Yu, Zollh{\"o}fer, and Rhodin]{su2021nerf}
Shih-Yang Su, Frank Yu, Michael Zollh{\"o}fer, and Helge Rhodin.
\newblock A-nerf: Articulated neural radiance fields for learning human shape,
  appearance, and pose.
\newblock \emph{Advances in Neural Information Processing Systems},
  34:\penalty0 12278--12291, 2021.

\bibitem[Su et~al.(2020)Su, Xu, Zheng, Yu, Liu, and Fang]{su2020robustfusion}
Zhuo Su, Lan Xu, Zerong Zheng, Tao Yu, Yebin Liu, and Lu Fang.
\newblock Robustfusion: Human volumetric capture with data-driven visual cues
  using a rgbd camera.
\newblock In \emph{Computer Vision--ECCV 2020: 16th European Conference,
  Glasgow, UK, August 23--28, 2020, Proceedings, Part IV 16}, pages 246--264.
  Springer, 2020.

\bibitem[Sun et~al.(2022)Sun, Sun, and Chen]{sun2022direct}
Cheng Sun, Min Sun, and Hwann-Tzong Chen.
\newblock Direct voxel grid optimization: Super-fast convergence for radiance
  fields reconstruction.
\newblock In \emph{Proceedings of the IEEE/CVF Conference on Computer Vision
  and Pattern Recognition}, pages 5459--5469, 2022.

\bibitem[Takikawa et~al.(2021)Takikawa, Litalien, Yin, Kreis, Loop,
  Nowrouzezahrai, Jacobson, McGuire, and Fidler]{takikawa2021neural}
Towaki Takikawa, Joey Litalien, Kangxue Yin, Karsten Kreis, Charles Loop, Derek
  Nowrouzezahrai, Alec Jacobson, Morgan McGuire, and Sanja Fidler.
\newblock Neural geometric level of detail: Real-time rendering with implicit
  3d shapes.
\newblock In \emph{Proceedings of the IEEE/CVF Conference on Computer Vision
  and Pattern Recognition}, pages 11358--11367, 2021.

\bibitem[Tewari et~al.(2022)Tewari, Thies, Mildenhall, Srinivasan, Tretschk,
  Yifan, Lassner, Sitzmann, Martin-Brualla, Lombardi,
  et~al.]{tewari2022advances}
Ayush Tewari, Justus Thies, Ben Mildenhall, Pratul Srinivasan, Edgar Tretschk,
  Wang Yifan, Christoph Lassner, Vincent Sitzmann, Ricardo Martin-Brualla,
  Stephen Lombardi, et~al.
\newblock Advances in neural rendering.
\newblock In \emph{Computer Graphics Forum}, pages 703--735. Wiley Online
  Library, 2022.

\bibitem[Thies et~al.(2019)Thies, Zollh{\"o}fer, and
  Nie{\ss}ner]{thies2019deferred}
Justus Thies, Michael Zollh{\"o}fer, and Matthias Nie{\ss}ner.
\newblock Deferred neural rendering: Image synthesis using neural textures.
\newblock \emph{Acm Transactions on Graphics (TOG)}, 38\penalty0 (4):\penalty0
  1--12, 2019.

\bibitem[Tiwari et~al.(2021)Tiwari, Sarafianos, Tung, and
  Pons-Moll]{tiwari2021neural}
Garvita Tiwari, Nikolaos Sarafianos, Tony Tung, and Gerard Pons-Moll.
\newblock Neural-gif: Neural generalized implicit functions for animating
  people in clothing.
\newblock In \emph{Proceedings of the IEEE/CVF International Conference on
  Computer Vision}, pages 11708--11718, 2021.

\bibitem[Trevithick and Yang(2021)]{trevithick2021grf}
Alex Trevithick and Bo Yang.
\newblock Grf: Learning a general radiance field for 3d representation and
  rendering.
\newblock In \emph{Proceedings of the IEEE/CVF International Conference on
  Computer Vision}, pages 15182--15192, 2021.

\bibitem[Wang et~al.(2022{\natexlab{a}})Wang, Wang, Sun, Kortylewski, and
  Yuille]{wang2022voge}
Angtian Wang, Peng Wang, Jian Sun, Adam Kortylewski, and Alan Yuille.
\newblock Voge: a differentiable volume renderer using gaussian ellipsoids for
  analysis-by-synthesis.
\newblock In \emph{The Eleventh International Conference on Learning
  Representations}, 2022{\natexlab{a}}.

\bibitem[Wang et~al.(2021{\natexlab{a}})Wang, Liu, Liu, Theobalt, Komura, and
  Wang]{wang2021neus}
Peng Wang, Lingjie Liu, Yuan Liu, Christian Theobalt, Taku Komura, and Wenping
  Wang.
\newblock Neus: Learning neural implicit surfaces by volume rendering for
  multi-view reconstruction.
\newblock \emph{arXiv preprint arXiv:2106.10689}, 2021{\natexlab{a}}.

\bibitem[Wang et~al.(2021{\natexlab{b}})Wang, Wang, Genova, Srinivasan, Zhou,
  Barron, Martin-Brualla, Snavely, and Funkhouser]{wang2021ibrnet}
Qianqian Wang, Zhicheng Wang, Kyle Genova, Pratul~P Srinivasan, Howard Zhou,
  Jonathan~T Barron, Ricardo Martin-Brualla, Noah Snavely, and Thomas
  Funkhouser.
\newblock Ibrnet: Learning multi-view image-based rendering.
\newblock In \emph{Proceedings of the IEEE/CVF Conference on Computer Vision
  and Pattern Recognition}, pages 4690--4699, 2021{\natexlab{b}}.

\bibitem[Wang et~al.(2022{\natexlab{b}})Wang, Schwarz, Geiger, and
  Tang]{wang2022arah}
Shaofei Wang, Katja Schwarz, Andreas Geiger, and Siyu Tang.
\newblock Arah: Animatable volume rendering of articulated human sdfs.
\newblock In \emph{European conference on computer vision}, 2022{\natexlab{b}}.

\bibitem[Wang et~al.(2004)Wang, Bovik, Sheikh, and Simoncelli]{wang2004image}
Zhou Wang, Alan~C Bovik, Hamid~R Sheikh, and Eero~P Simoncelli.
\newblock Image quality assessment: from error visibility to structural
  similarity.
\newblock \emph{IEEE transactions on image processing}, 13\penalty0
  (4):\penalty0 600--612, 2004.

\bibitem[Weng et~al.(2020)Weng, Curless, and
  Kemelmacher-Shlizerman]{weng2020vid2actor}
Chung-Yi Weng, Brian Curless, and Ira Kemelmacher-Shlizerman.
\newblock Vid2actor: Free-viewpoint animatable person synthesis from video in
  the wild.
\newblock \emph{arXiv preprint arXiv:2012.12884}, 2020.

\bibitem[Weng et~al.(2022)Weng, Curless, Srinivasan, Barron, and
  Kemelmacher-Shlizerman]{weng_humannerf_2022_cvpr}
Chung-Yi Weng, Brian Curless, Pratul~P. Srinivasan, Jonathan~T. Barron, and Ira
  Kemelmacher-Shlizerman.
\newblock Human{N}e{RF}: Free-viewpoint rendering of moving people from
  monocular video.
\newblock In \emph{Proceedings of the IEEE/CVF Conference on Computer Vision
  and Pattern Recognition (CVPR)}, pages 16210--16220, 2022.

\bibitem[Weng et~al.(2023)Weng, Wang, and Yeung]{weng2023zeroavatar}
Zhenzhen Weng, Zeyu Wang, and Serena Yeung.
\newblock Zeroavatar: Zero-shot 3d avatar generation from a single image.
\newblock \emph{arXiv preprint arXiv:2305.16411}, 2023.

\bibitem[Wigner(1931)]{wigner1931gruppentheorie}
EP Wigner.
\newblock Gruppentheorie und ihre anwendungen auf die quantenmechanik der
  atomspektren. friedr. vieweg und sohn akt.
\newblock \emph{Ges., Braunschweig}, 1931.

\bibitem[Wiles et~al.(2020)Wiles, Gkioxari, Szeliski, and
  Johnson]{wiles2020synsin}
Olivia Wiles, Georgia Gkioxari, Richard Szeliski, and Justin Johnson.
\newblock Synsin: End-to-end view synthesis from a single image.
\newblock In \emph{Proceedings of the IEEE/CVF Conference on Computer Vision
  and Pattern Recognition}, pages 7467--7477, 2020.

\bibitem[Wu et~al.(2023)Wu, Yi, Fang, Xie, Zhang, Wei, Liu, Tian, and
  Wang]{wu20234d}
Guanjun Wu, Taoran Yi, Jiemin Fang, Lingxi Xie, Xiaopeng Zhang, Wei Wei, Wenyu
  Liu, Qi Tian, and Xinggang Wang.
\newblock 4d gaussian splatting for real-time dynamic scene rendering.
\newblock \emph{arXiv preprint arXiv:2310.08528}, 2023.

\bibitem[Xu et~al.(2021)Xu, Alldieck, and Sminchisescu]{xu2021h}
Hongyi Xu, Thiemo Alldieck, and Cristian Sminchisescu.
\newblock H-nerf: Neural radiance fields for rendering and temporal
  reconstruction of humans in motion.
\newblock \emph{Advances in Neural Information Processing Systems}, 34, 2021.

\bibitem[Xu et~al.(2022)Xu, Xu, Philip, Bi, Shu, Sunkavalli, and
  Neumann]{xu2022point}
Qiangeng Xu, Zexiang Xu, Julien Philip, Sai Bi, Zhixin Shu, Kalyan Sunkavalli,
  and Ulrich Neumann.
\newblock Point-nerf: Point-based neural radiance fields.
\newblock In \emph{Proceedings of the IEEE/CVF Conference on Computer Vision
  and Pattern Recognition}, pages 5438--5448, 2022.

\bibitem[Xu et~al.(2023)Xu, Peng, Lin, He, Sun, Shen, Bao, and
  Zhou]{xu20234k4d}
Zhen Xu, Sida Peng, Haotong Lin, Guangzhao He, Jiaming Sun, Yujun Shen, Hujun
  Bao, and Xiaowei Zhou.
\newblock 4k4d: Real-time 4d view synthesis at 4k resolution.
\newblock \emph{arXiv preprint arXiv:2310.11448}, 2023.

\bibitem[Yang et~al.(2020)Yang, Mao, Alvarez, and Liu]{yang2020cost}
Jiayu Yang, Wei Mao, Jose~M Alvarez, and Miaomiao Liu.
\newblock Cost volume pyramid based depth inference for multi-view stereo.
\newblock In \emph{Proceedings of the IEEE/CVF Conference on Computer Vision
  and Pattern Recognition}, pages 4877--4886, 2020.

\bibitem[Yang et~al.(2021)Yang, Wang, Manivasagam, Huang, Ma, Yan, Yumer, and
  Urtasun]{yang2021s3}
Ze Yang, Shenlong Wang, Sivabalan Manivasagam, Zeng Huang, Wei-Chiu Ma, Xinchen
  Yan, Ersin Yumer, and Raquel Urtasun.
\newblock S3: Neural shape, skeleton, and skinning fields for 3d human
  modeling.
\newblock In \emph{Proceedings of the IEEE/CVF conference on computer vision
  and pattern recognition}, pages 13284--13293, 2021.

\bibitem[Yang et~al.(2023{\natexlab{a}})Yang, Gao, Zhou, Jiao, Zhang, and
  Jin]{yang2023deformable}
Ziyi Yang, Xinyu Gao, Wen Zhou, Shaohui Jiao, Yuqing Zhang, and Xiaogang Jin.
\newblock Deformable 3d gaussians for high-fidelity monocular dynamic scene
  reconstruction.
\newblock \emph{arXiv preprint arXiv:2309.13101}, 2023{\natexlab{a}}.

\bibitem[Yang et~al.(2023{\natexlab{b}})Yang, Yang, Pan, Zhu, and
  Zhang]{yang2023real}
Zeyu Yang, Hongye Yang, Zijie Pan, Xiatian Zhu, and Li Zhang.
\newblock Real-time photorealistic dynamic scene representation and rendering
  with 4d gaussian splatting.
\newblock \emph{arXiv preprint arXiv:2310.10642}, 2023{\natexlab{b}}.

\bibitem[Yifan et~al.(2019)Yifan, Serena, Wu, {\"O}ztireli, and
  Sorkine-Hornung]{yifan2019differentiable}
Wang Yifan, Felice Serena, Shihao Wu, Cengiz {\"O}ztireli, and Olga
  Sorkine-Hornung.
\newblock Differentiable surface splatting for point-based geometry processing.
\newblock \emph{ACM Transactions on Graphics (TOG)}, 38\penalty0 (6):\penalty0
  1--14, 2019.

\bibitem[Yu et~al.(2021{\natexlab{a}})Yu, Li, Tancik, Li, Ng, and
  Kanazawa]{yu2021plenoctrees}
Alex Yu, Ruilong Li, Matthew Tancik, Hao Li, Ren Ng, and Angjoo Kanazawa.
\newblock Plenoctrees for real-time rendering of neural radiance fields.
\newblock In \emph{Proceedings of the IEEE/CVF International Conference on
  Computer Vision}, pages 5752--5761, 2021{\natexlab{a}}.

\bibitem[Yu et~al.(2021{\natexlab{b}})Yu, Ye, Tancik, and
  Kanazawa]{yu2021pixelnerf}
Alex Yu, Vickie Ye, Matthew Tancik, and Angjoo Kanazawa.
\newblock {pixelNeRF}: Neural radiance fields from one or few images.
\newblock In \emph{CVPR}, 2021{\natexlab{b}}.

\bibitem[Zhang et~al.(2020)Zhang, Riegler, Snavely, and
  Koltun]{zhang2020nerf++}
Kai Zhang, Gernot Riegler, Noah Snavely, and Vladlen Koltun.
\newblock Nerf++: Analyzing and improving neural radiance fields.
\newblock \emph{arXiv preprint arXiv:2010.07492}, 2020.

\bibitem[Zhang et~al.(2022)Zhang, Baek, Rusinkiewicz, and
  Heide]{zhang2022differentiable}
Qiang Zhang, Seung-Hwan Baek, Szymon Rusinkiewicz, and Felix Heide.
\newblock Differentiable point-based radiance fields for efficient view
  synthesis.
\newblock In \emph{SIGGRAPH Asia 2022 Conference Papers}, pages 1--12, 2022.

\bibitem[Zhang et~al.(2018)Zhang, Isola, Efros, Shechtman, and
  Wang]{zhang2018unreasonable}
Richard Zhang, Phillip Isola, Alexei~A Efros, Eli Shechtman, and Oliver Wang.
\newblock The unreasonable effectiveness of deep features as a perceptual
  metric.
\newblock In \emph{Proceedings of the IEEE conference on computer vision and
  pattern recognition}, pages 586--595, 2018.

\bibitem[Zhao et~al.(2021)Zhao, Yang, Zhang, Lin, Zhang, Yu, and
  Xu]{zhao2021humannerf}
Fuqiang Zhao, Wei Yang, Jiakai Zhang, Pei Lin, Yingliang Zhang, Jingyi Yu, and
  Lan Xu.
\newblock Humannerf: Generalizable neural human radiance field from sparse
  inputs.
\newblock \emph{arXiv preprint arXiv:2112.02789}, 3, 2021.

\bibitem[Zhou et~al.(2016)Zhou, Tulsiani, Sun, Malik, and Efros]{zhou2016view}
Tinghui Zhou, Shubham Tulsiani, Weilun Sun, Jitendra Malik, and Alexei~A Efros.
\newblock View synthesis by appearance flow.
\newblock In \emph{Computer Vision--ECCV 2016: 14th European Conference,
  Amsterdam, The Netherlands, October 11--14, 2016, Proceedings, Part IV 14},
  pages 286--301. Springer, 2016.

\bibitem[Zwicker et~al.(2001{\natexlab{a}})Zwicker, Pfister, Van~Baar, and
  Gross]{zwicker2001ewa}
Matthias Zwicker, Hanspeter Pfister, Jeroen Van~Baar, and Markus Gross.
\newblock Ewa volume splatting.
\newblock In \emph{Proceedings Visualization, 2001. VIS'01.}, pages 29--538.
  IEEE, 2001{\natexlab{a}}.

\bibitem[Zwicker et~al.(2001{\natexlab{b}})Zwicker, Pfister, Van~Baar, and
  Gross]{zwicker2001surface}
Matthias Zwicker, Hanspeter Pfister, Jeroen Van~Baar, and Markus Gross.
\newblock Surface splatting.
\newblock In \emph{Proceedings of the 28th annual conference on Computer
  graphics and interactive techniques}, pages 371--378, 2001{\natexlab{b}}.

\end{thebibliography}
}

\clearpage
\appendix

\section{Implementation Details}

\noindent \textbf{Implementation Details of \nickname{}} 
Our pose refinement module consists of 4 fully connected layers, \ie, an input layer, 2 hidden layers, and one output layer.
Each layer is followed by a ReLU activation. 
The dimension of the hidden layer is 128, while the input and output dimension of the pose refinement module is 69.
The LBS offset module adopts 5 fully connected layers with an input layer, 3 hidden layers, and one output layer.
The ReLU activation is used after each layer.
Positional encoding is applied to the 3D Gaussian positions before they are fed into the input layer.
The input and output dimension of the LBS offset module is 63 and 24 (number of joints).
We use Adam optimizer with a learning rate $10^{-5}$ to optimize the above two modules. 
We set the threshold of KL divergence as 0.4 to perform split/clone operations.
Other training details for 3D Gaussians are the same as~\cite{kerbl20233d}.

\noindent \textbf{Evaluation Metrics.}
To quantitatively evaluate the quality of rendered novel view and novel pose images, we report the peak signal-to-noise ratio (PSNR) ~\cite{sara2019image}, structural similarity index (SSIM)~\cite{wang2004image} and Learned Perceptual Image Patch Similarity (LPIPS)~\cite{zhang2018unreasonable}.

\noindent \textbf{Details of Comparable Methods.} 1). Subject-specific optimization-based methods. 
Neural Body (NB)~\cite{neuralbody} encodes latent codes in SMPL vertex points and uses them to learn the neural radiance fields.
Animatable NeRF (AN)~\cite{peng2021animatable} learns a canonical human NeRF through skeleton-driven deformation and learned blend weight fields.
AS\cite{peng2022animatable} further extends~\cite{peng2021animatable} by learning a signed distance field and a pose-dependent deformation field for residual information and geometric details of dynamic 3D humans.
HumanNeRF~\cite{weng_humannerf_2022_cvpr} incorporates a pose refinement module, LBS field, and non-rigid deformation module to optimize a volumetric representation of 3D humans in the canonical space.
Unlike the above methods using coordinated neural networks to learn neural radiance fields, InstantNVR~\cite{geng2023learning} and InstantAvatar~\cite{jiang2023instantavatar} propose to use multi-hashing encoding for fast training of 3D humans. 
Given the fast rendering speed of mixtures
of volumetric primitives~\cite{lombardi2021mixture}, DVA further extends it to articulated 3D humans for high-quality telepresence. 
Specifically, DVA first attaches a fixed number (4096) of voxels in the UV space and then transforms voxels to posed space with the sparse multi-view input images as the conditioning information, along with an MLP decoder to regress the opacity and color information.
2). Generalizable methods. PixelNeRF~\cite{yu2021pixelnerf} learns a neural network to infer the radiance field based on the input image. 
Neural Human Performer (NHP)~\cite{kwon2021neural} aggregates pixel-aligned features at each time step and temporally-fused features to learn generalizable neural radiance fields. 
For generalizable methods, we evaluate each subject (\eg, one subject of MonoCap) by first pre-training the model on the other data set (e.g., ZJU\_Mocap data set) and then fine-tuning it on the evaluated subject.

\noindent \textbf{Efficient Implementation of KL Divergence}
As the covariance matrix is decomposed into the product of rotation and scaling matrices $\bm{\Sigma}=\bm{R}\bm{S}\bm{S}^{T}\bm{R}^{T}$, we simplify the computation of matrix inverse and determinant operations, \ie, 
\begin{equation}
\begin{aligned}
\bm{\Sigma}_1^{-1} &= (\bm{R}\bm{S}\bm{S}^{T}\bm{R}^{T})^{-1}=\bm{R}\bm{S}^{-1}\bm{S}^{-1}\bm{R}^{T},\\
\det \bm{\Sigma}_1 &= \det(\bm{R}\bm{S}\bm{S}^{T}\bm{R}^{T})= \det(\bm{S}) * \det(\bm{S})
\end{aligned}
\end{equation} 
Since scaling matrix $\bm{S}$ is a diagonal matrix, the inverse and determinant of a diagonal matrix can be easily derived by inversing and prodding the diagonal elements respectively. 
Meanwhile, the inverse of the orthogonal rotation matrix is the transpose of the original matrix.
The above simplification saves the computation time for matrix inverse and determinant operation. 

\noindent \textbf{Rotating Spherical Harmonic coefficients} When transforming 3D Gaussians from canonical space to posed space, the SH coefficients should also be rotated for view-dependent color effects.
The above is achieved by first computing a Wigner D-matrix~\cite{wigner1931gruppentheorie} and then rotating SH coefficients with the Wigner D-matrix.
In our implementation, we find that rotating SH coefficients has little effect on the final performance\footnote{We also find that the implementation of Wigner D-matrix using Pytorch~\cite{paszke2019pytorch} is time-consuming due to the matrix exponential operation.}, so we do not consider it in our work.

\section{Details of Loss Functions}
\noindent \textbf{Photometric Loss.} 
Given the ground truth target image $C$  and predicted image $\hat{C}$, we apply the photometric loss:
\begin{equation}
\label{loss: color}
\begin{aligned}
\mathcal{L}_{color} =||\hat{C} - C||_2.
\end{aligned}
\end{equation}

\noindent \textbf{Mask Loss.} 
We also leverage the human region masks for Human NeRF optimization. The mask loss is defined as:
\begin{equation}
\label{loss: mask}
\begin{aligned}
\mathcal{L}_{mask} = ||\hat{M} - M||_2,
\end{aligned}
\end{equation}
where $\hat{M}$ is the accumulated volume density and $M$ is the ground truth binary mask label.

\noindent \textbf{SSIM Loss.} 
We further employ SSIM to ensure the structural similarity between ground truth and synthesized images:
\begin{equation}
\label{loss: ssim}
\begin{aligned}
\mathcal{L}_{SSIM} = \text{SSIM}(\hat{C}, C).
\end{aligned}
\end{equation}

\noindent \textbf{LPIPS Loss.}
The perceptual loss LPIPS is also utilized to ensure the quality of rendered image, \ie,
\begin{equation}
\label{loss: lpips}
\begin{aligned}
\mathcal{L}_{LPIPS} = \text{LPIPS}(\hat{C}, C).
\end{aligned}
\end{equation}

\section{Experiments on DNA-Rendering}
We further evaluate the performance of our \nickname{} and two representative baseline methods on a DNA-Rendering data set.
We select two sequences (0012\_09 and 0025\_11 from part 1) from the DNA-Rendering data set and collect 100 frames for training.
Similar to ZJU\_MoCap and MonoCap, one camera is used for training.
For evaluation purposes, we use four nearby camera views as testing views.
As shown in Tab.~\ref{tab: sup_dna_rendering_result} and Fig, AS~\cite{peng2022animatable} and InstantAvatar~\cite{jiang2023instantavatar} struggle to produce photorealistic renderings due to the complex clothing and fast-moving human actors recorded on the DNA-Rendering data set.
In comparison, our \nickname{} learns high-quality 3D human performers with fast training and rendering speed, which verifies the flexibility and efficiency of 3D Gaussian Splatting.

\begin{table}[h]
\setlength{\abovecaptionskip}{0cm}
\vspace{-4mm}
\caption{Quantitative comparison of our \nickname{} and baseline methods on the DNA-Rendering data set. LPIPS$^*$ = 1000 $\times$ LPIPS. Frames per second (FPS) are measured on an RTX 3090.
}
\centering
\label{tab: sup_dna_rendering_result}
\setlength{\tabcolsep}{1.0mm}{
\begin{tabular}{l|ccccc}
\toprule
\multirow{2}*{Method} & \multicolumn{5}{c}{DNA-Rendering} \\
~ & PSNR$\uparrow$ & SSIM$\uparrow$ & LPIPS$^*$$\downarrow$ & Train & FPS \\
\midrule
AS~\cite{peng2022animatable} & 27.67 & 0.954 & 50.99 & 10h & 0.14 \\
InstantAvatar~\cite{jiang2023instantavatar} & 24.77 & 0.922 & 78.55 & 20m & 0.48 \\
\textbf{\nickname{}}(Ours) & \textbf{29.11} & \textbf{0.961} & \textbf{37.68} & \textbf{4m} & \textbf{152} \\
\bottomrule
\end{tabular}}
\vspace{-4mm}
\end{table}

\begin{figure}[t]
    \vspace{-3mm}
    \setlength{\abovecaptionskip}{0cm}
    \centering
    \includegraphics[width=8.5cm]{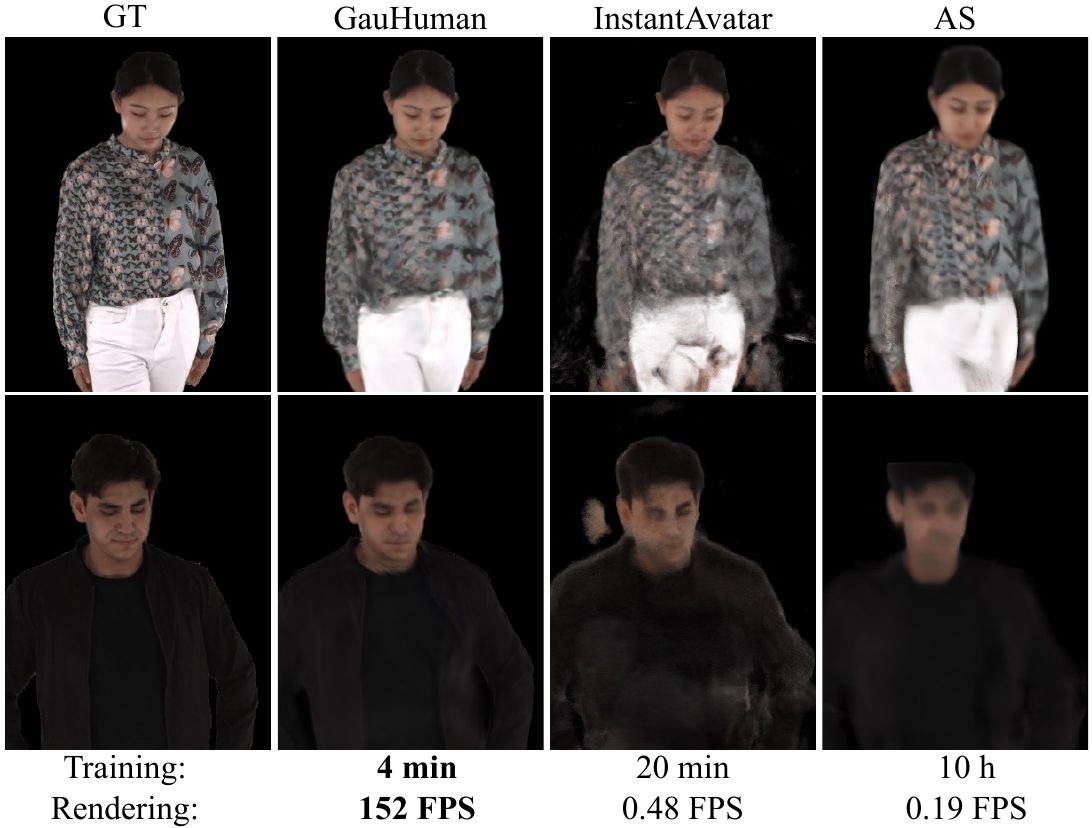}
    \setlength{\abovecaptionskip}{0cm}
    \caption{Novel view synthesis results produced by our \nickname{} and baseline methods on DNA Rendering data set. The bottom lines show the training time and rendering speed of each method on the DNA Rendering data set. Zoom in for the best view.} 
\label{fig: sup_vis}
\vspace{-3mm}
\end{figure}

\section{Comparison with concurrent work GART~\cite{lei2023gart}}
Our concurrent work GART~\cite{lei2023gart} also extends Gaussian Splatting to 3D human modelling with monocular videos. 
It achieves comparable novel view synthesis performance when compared with state-of-the-art baseline methods on ZJU\_MoCap data set while improving the rendering speed to 77 FPS with the efficient 3D Gaussian Splatting technique.
We reproduce the result of GART with their released code and show the comparison results in Tab.~\ref{tab: supp_com_gauhuman_gart}.
For a fair comparison, we do not conduct test-time optimization of SMPL parameters with images from the test set on GART~\cite{lei2023gart}.
In comparison with GART, our \nickname{} produces slightly better novel view synthesis performance with both faster training (\textit{1m vs. 3m}) and rendering (\textit{189FPS vs. 77FPS}) speed.
Specifically, we achieve fast optimization of GauHuman by initializing and pruning 3D Gaussians with 3D human prior, while splitting/cloning via KL divergence guidance, along with a novel merge operation for further speeding up.
Notably, without sacrificing rendering quality, \nickname{} can fast model the 3D human performer with $\sim$13k 3D Gaussians.

\begin{table}[h]
\vspace{-3mm}
\setlength{\abovecaptionskip}{0cm}
\caption{Quantitative comparison of our \nickname{} and GART on the ZJU\_MoCap data set. LPIPS$^*$ = 1000 $\times$ LPIPS. Frames per second (FPS) is measured on an RTX 3090. \textbf{For a fair comparison, we do not conduct test-time optimization of SMPL parameters with images from the test set on GART~\cite{lei2023gart}.}
}
\small
\centering
\label{tab: supp_com_gauhuman_gart}
\setlength{\tabcolsep}{1.0mm}{
\begin{tabular}{lcccccc}
\toprule
\multirow{2}*{Method} & \multicolumn{6}{c}{ZJU\_MoCap (Avg)}\\
 & PSNR$\uparrow$ & SSIM$\uparrow$ & LPIPS$^*$$\downarrow$ & \#Gau & Train & FPS\\
\midrule
\rowcolor{black!15}
GART & 30.91 & 0.9615 & 31.83 & 53.4k & 3m & 77\\
\rowcolor{black!15}
GauHuman (Ours) & \textbf{31.34} & \textbf{0.9647} & \textbf{30.51} & \textbf{11.8k} & \textbf{1m} & \textbf{189} \\
\bottomrule
 & \multicolumn{4}{c}{my\_377}\\
 & PSNR$\uparrow$ & SSIM$\uparrow$ & LPIPS$^*$$\downarrow$ & \#Gau \\
\midrule
GART & 31.90 & 0.9747 & \textbf{18.8} & 55.0k \\
GauHuman (Ours) & \textbf{32.24} & \textbf{0.9757} & 18.9 & \textbf{12.6k}\\
\bottomrule
 & \multicolumn{4}{c}{my\_386}\\
 & PSNR$\uparrow$ & SSIM$\uparrow$ & LPIPS$^*$$\downarrow$ & \#Gau \\
\midrule
GART & 33.50 & 0.9669 & 29.9 & 51.4k\\
GauHuman (Ours) & \textbf{33.72} & \textbf{0.9693} & \textbf{29.0} & \textbf{13.1k} \\
\bottomrule
 & \multicolumn{4}{c}{my\_387}\\
 & PSNR$\uparrow$ & SSIM$\uparrow$ & LPIPS$^*$$\downarrow$ & \#Gau \\
\midrule
GART & 27.74 & 0.9518 & 40.3 & 52.9k\\
GauHuman (Ours) & \textbf{28.19} & \textbf{0.9564} & \textbf{39.3} & \textbf{9.9k}\\
\bottomrule
 & \multicolumn{4}{c}{my\_392}\\
 & PSNR$\uparrow$ & SSIM$\uparrow$ & LPIPS$^*$$\downarrow$ & \#Gau \\
\midrule
GART & 31.92 & 0.9637 & 32.6 & 51.6k\\
GauHuman (Ours) & \textbf{32.27} & \textbf{0.9669} & \textbf{30.2} & \textbf{11.3k}\\
\bottomrule
 & \multicolumn{4}{c}{my\_393}\\
 & PSNR$\uparrow$ & SSIM$\uparrow$ & LPIPS$^*$$\downarrow$ & \#Gau \\
\midrule
GART & 29.34 & 0.9540 & 37.9 & 51.7k\\
GauHuman (Ours) & \textbf{30.24} & \textbf{0.9584} & \textbf{35.2} & \textbf{11.0k}\\
\bottomrule
 & \multicolumn{4}{c}{my\_394}\\
 & PSNR$\uparrow$ & SSIM$\uparrow$ & LPIPS$^*$$\downarrow$ & \#Gau \\
\midrule
GART & 31.08 & 0.9577 & 31.5 & 57.7k\\
GauHuman (Ours) & \textbf{31.42} & \textbf{0.9611} & \textbf{30.6} & \textbf{12.8k}\\
\bottomrule
\end{tabular}}
\vspace{-3mm}
\end{table}

\end{document}